\documentclass[final,1p,times,authoryear]{elsarticle}

\usepackage[table,xcdraw]{xcolor}

\usepackage{amssymb}
\usepackage{amsmath}

\usepackage{longtable}

\usepackage{booktabs}

\usepackage{array}
\usepackage{geometry}
\definecolor{grey}{rgb}{0.5,0.5,0.5}
\usepackage{lineno}
\usepackage{comment}
\usepackage[inline]{enumitem}
\usepackage[most]{tcolorbox}
\definecolor{lightgray}{gray}{0.93}
\definecolor{secondarygray}{gray}{0.98}
\usepackage{booktabs}
\usepackage{multirow}
\usepackage{array}

\newcolumntype{L}[1]{>{\raggedright\arraybackslash}p{#1}}
\newcolumntype{C}[1]{>{\centering\arraybackslash}p{#1}}
\usepackage{colortbl} 
\usepackage{svg}
\usepackage[colorlinks=true, citecolor=blue, linkcolor=black, urlcolor=blue]{hyperref}
\newcommand{\adil}[1]{\textcolor{purple}{Adil:~#1}}
\newcommand{\michael}[1]{\textcolor{olive}{Michael:~#1}}

\newcommand{\franz}[1]{\textcolor{green}{Franz:~#1}}
\newcommand{\change}[2]{\textcolor{red}{#2}}

\journal{Energy and AI}

\begin{document}

\newcommand{\RQOne}{\textbf{RQ1} : To what extent do ML-based FDD studies for HVAC systems provide complete methodological information such as datasets description, data pre-processing, hyperparameter tuning and model training within their manuscripts?
}

\newcommand{\RQTwo}{\textbf{RQ2} : What percentage of studies make key artifacts (code, datasets, trained models) available through external sources (e.g., GitHub, institutional repositories)?
}

\newcommand{\RQThree}{\textbf{RQ3} : To what extent do studies report data splitting schemes, evaluation strategies, and performance metrics?
}

\newcommand{\RQSecondary}{\textbf{What insights can bibliometric indicators—such as citation count, h-index, or publication history—offer regarding artifact-sharing practices in ML-based FDD for HVAC systems studies?}}

\newcommand{\DimensionOne}{\textbf{$D_1$ (Data)}: Datasets, along with their types and descriptions, serve as essential references for reproducing results and conducting experiments.}
\newcommand{\DimensionTwo}{\textbf{$D_2$ (Method)}: The information related to data pre-processing steps, feature engineering techniques, and model development procedures along with the coding scripts is essential for methodological and results reproducibility.}

\newcommand{\DimensionThree}{\textbf{$D_3$ (Experiment)}: To support reproducible experiments and results, comprehensive documentation and the sharing of code package related to data, optimal methods, evaluation procedures, and metrics are essential.}
\begin{frontmatter}

\title{Reproducibility of Machine Learning-Based Fault Detection and Diagnosis for HVAC Systems in Buildings: An Empirical Study}

\author[1] {Adil Mukhtar}
\ead{adil.mukhtar@tuwien.ac.at}

\author[1]{Michael Hadwiger}
\ead{michael.hadwiger@tuwien.ac.at}

\author[2]{Franz Wotawa}
\ead{wotawa@tugraz.at}

\author[1]{Gerald Schweiger}
\ead{gerald.schweiger@tuwien.ac.at}

\affiliation[1]{organization={TU Wien}, city={Vienna}, country={Austria}}

\affiliation[2]{organization={Graz University of Technology}, city={Graz}, country={Austria}}

\begin{abstract}
Reproducibility is a cornerstone of scientific research, enabling independent verification and validation of empirical findings. 
The topic gained prominence in fields such as psychology and medicine, where concerns about non-replicable results sparked ongoing discussions about research practices. 
In recent years, the fast-growing field of Machine Learning (ML) has become part of this discourse, as it faces similar concerns about transparency and reliability.
Some reproducibility issues in ML research are shared with other fields, such as limited access to data and missing methodological details. 
In addition, ML introduces specific challenges, including inherent nondeterminism and computational constraints.
While reproducibility issues are increasingly recognized by the ML community and its major conferences, less is known about how these challenges manifest in applied disciplines. 
This paper contributes to closing this gap by analyzing the transparency and reproducibility standards of ML applications in building energy systems.
The results indicate that nearly all articles are not reproducible due to insufficient disclosure across key dimensions of reproducibility. 72\% of the articles do not specify whether the dataset used is public, proprietary, or commercially available. 
Only two papers share a link to their code—one of which was broken.
Two-thirds of the publications were authored exclusively by academic researchers, yet no significant differences in reproducibility were observed compared to publications with industry-affiliated authors.
These findings highlight the need for targeted interventions, including reproducibility guidelines, training for researchers, and policies by journals and conferences that promote transparency and reproducibility. 

\end{abstract}





\end{frontmatter}




\section{Introduction}
\label{sec:introduction}
Buildings account for almost 40\% of global energy use and contribute approximately 20\% to global carbon emissions~\citep{schweiger2020active}. This significant share underscores the critical role of the building sector in environmental sustainability and in achieving global energy efficiency. The timely advancement of, perhaps serendipitously aligned with growing environmental concerns, Internet of Things (IoT) technologies has not only guided the evolution of building development towards the concept of \emph{smart buildings}~\citep{snoonian2003smart} but also supported and accelerated their growth~\citep{jia2019adopting}. Smart buildings, in the context of Information and Communication Technology (ICT), are defined as cyber-physical systems that enable intelligent decision-making through continuous monitoring and control of building operations. This is achieved through real-time data acquisition and communication facilitated by an IoT layer integrated into the building system architecture. Like any other complex system, building systems are prone to faults, which can lead to undesirable outcomes such as increased energy waste, compromised occupant comfort, and high maintenance costs. 

Over the years, many approaches have been proposed to detect defective states and identify potential causes in various components, primarily Heating, Ventilation, and Air Conditioning (HVAC) systems. These methods are commonly referred to as fault detection and diagnosis (FDD) or, synonymously, automated fault detection and diagnosis (AFDD)~\citep{katipamula2005methodsI, katipamula2005methodsII, chen2022review}. 
In the context of smart buildings, for example, faults can arise from various source such as sensor failures, control errors, or equipment degradation each leading to distinct performance issues. A fault in an HVAC system can result in excessive energy consumption, reduced thermal comfort, or even system failure. Depending on the type and severity of the fault, different detection and diagnostic strategies may be required. Early detection and diagnosis of faults is crucial, as studies have shown that operational faults account for approximately 15–30\% of energy losses in commercial buildings~\citep{nelson2022machine}. 

\begin{figure}[htbp]
    \centering
    \includegraphics[clip, trim=0.0cm 1.5cm 0.1cm 0.5cm, width=1\textwidth]{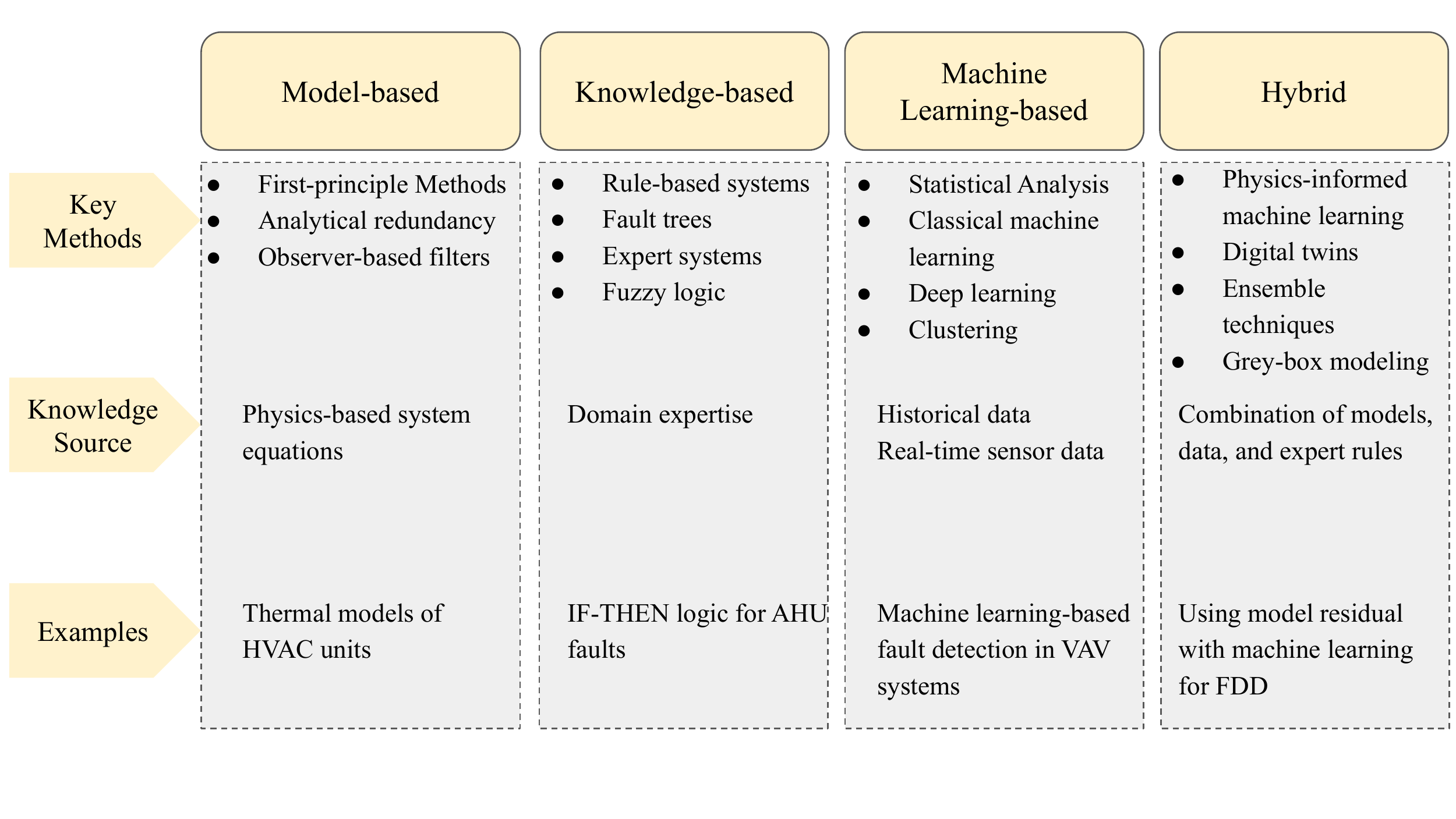}
    \caption{Overview of categorized FDD methods for building systems
    }
    \label{fig:fdd-types}
\end{figure}

Among the proposed FDD methodologies, some rely on \emph{model-based} approaches using mathematical representations of system dynamics, while others adopt \emph{knowledge-based} strategies guided by expert rules and predefined fault signatures~\citep{katipamula2005methodsI, katipamula2005methodsII}. However, more recently, \emph{machine learning-based} approaches have gained importance and became popular. These methods employ machine learning and statistical models to identify faults by analyzing large datasets from building automation systems (BAS). Proposals in this category range from traditional supervised learning techniques to advanced deep learning architectures capable of detecting subtle anomalies in high-dimensional space~\citep{matetic2022review, chen2023review}. While numerous studies have reviewed FDD methods for building systems, a summarized classification is provided in Figure~\ref{fig:fdd-types}. In the following, we briefly describe the motivation for conducting this study, along with the objectives and contributions of the study.

\subsection{Motivation and Background}
\label{sec:motivation-background}

Machine learning approaches have gained significant prominence in the field of building systems~\citep{matetic2022review, chen2023review, mirnaghi2020fault}. The terms \emph{data-driven} and \emph{machine learning} are often used interchangeably in the literature. In this article, we refer to FDD methods as data-driven if they perform fault detection and/or diagnosis by training a model on data or by employing statistical analysis techniques~\citep{chen2023review}. Nevertheless, these methods utilize historical operational data to learn patterns indicative of both normal and faulty behavior, employing algorithms such as artificial neural networks~\citep{jones2015fault}, support vector machines~\citep{namburu2007data}, Bayesian networks~\citep{xiao2014bayesian}, and various clustering techniques ~\citep{du2014fault, li2018improved, capozzoli2015fault}. More recently, researchers have investigated the application of advanced language models (e.g., GPT-3.5/4) for FDD and other energy services in buildings, highlighting both limitations and potential~\citep{zhang2025domain, liu2025large, langer2024large}.

Despite the growing adoption of ML methodologies and the increasing sophistication of ML models in recent research, we observed sustained shortcomings in the transparency of model development and evaluation procedures. For instance,~\cite{haibe2020transparency} highlighted even though~\citet{mckinney2020international} compellingly demonstrated the potential of AI techniques in advancing medical imaging, its failure to provide sufficiently detailed methodologies, complete coding artifacts, and openly accessible data repositories severely compromises its scientific credibility and erodes trust in its findings, despite the apparent promise of its results. Analogously,~\citet{popper2005logic} expressed this clearly: ``Non-reproducible single occurrences are of no significance for science". Reproducibility enables independent verification of results, the identification of errors, and the cumulative advancement of knowledge. Without it, research findings become epistemologically unreliable and ultimately call into question the progress of the entire field. In practice, this can undermine trust in scientific work — especially in experimental results — and prevent potentially valuable methods, such as those from the field of machine learning, from being applied~\citep{herrmann2024position}.

In general, \emph{reproducibility} refers to the ability to independently validate a study's proposed models and claims using the information, documentation, and supplementary materials provided in the manuscript or through externally accessible resources~\citep{gundersen2018state}. Hence, researchers are encouraged to share their research artifacts, including datasets, executable code, and detailed documentation, alongside clear descriptions of evaluation metrics, experimental procedures, and data partitioning strategies (e.g., train-test splits). Such transparency is essential for ensuring reproducibility and guarding against the risk of ``phantom progress''~\citep{ferrari2019we}, a term used to describe cases where models are insufficiently trained, evaluated on weak baselines, or presented with incomplete experimental details, thereby undermining independent validation and meaningful comparison.

\subsection{Terminologies \& Definitions}
A long-standing debate over the terms \emph{reproducibility} and \emph{replicability} across scientific disciplines has often led to greater confusion than a concrete resolution~\citep{drummond2009replicability, national2019reproducibility}. In efforts to distinguish between the two, researchers have proposed definitions grounded not in dictionary semantics, but in the practical and methodological requirements needed to verify the results claimed by a given study~\citep{drummond2009replicability, patil2016statistical, national2019reproducibility}. However, for the sake of clarity and to avoid engaging in debates over the nuanced distinction between the two terms, we adopt the definitions provided by~\citep{national2019reproducibility}. This allows us and the reader to maintain consistency and avoid ambiguity throughout this article. According to~\citep{national2019reproducibility}, the terms \emph{reproducibility} and \emph{replicability} are defined as follows:

``\emph{Reproducibility} is obtaining consistent results using the same input data; computation steps, methods, and code; and conditions of analysis."

``\emph{Replicability} is obtaining consistent results across studies aimed at answering the same scientific question, each of which has obtained its own data."

We also consider the domain specific definition for machine learning presented by ~\cite{gundersen2018state} relevant and incorporate through out this review:

``Reproducibility in empirical AI research is the
ability of an independent research team to produce the same
results using the same AI method based on the documentation made by the original research team."

This distinction is also supported by~\citet{nichols2021better}. However, in contrast to this view,~\citet{drummond2009replicability} argues that what is commonly referred to as “replicability” in the machine learning community—namely, the exact repetition of experiments through shared code and artifacts—should not be conflated with scientific reproducibility. According to Drummond, scientific reproducibility involves obtaining consistent results through different experiments and under varying conditions, thereby reflecting a more robust validation of scientific claims. His perspective, in our view, aligns reproducibility with the broader concept of methodological generalizability.

Despite these differing interpretations, our primary aim in presenting these definitions is to highlight the ongoing ambiguity surrounding the use of these terms across disciplines. 

\subsection{Objectives and Contribution}
\label{sec:objectives-contributions}

In this study, we investigate the state of reproducibility within the domain of building systems, with a particular focus on machine learning-based fault detection and diagnosis (FDD) in HVAC systems. To this purpose, we assess whether the field exhibits the same persistent reproducibility issues observed in other disciplines~\citep{haibe2020transparency}, such as insufficient artifact sharing and information on evaluation procedures, and unavailability of code utility packages. To systematically structure our investigation, we adopt and extend the reproducibility spectrum framework~\citep{peng2011reproducibility}—ranging from basic methodological transparency in publications to full computational replication.

We extend this framework to assess not only externally shared materials, but also the extent to which reproducibility-relevant information is documented within the manuscript itself. This adaptation addresses field-specific constraints while maintaining transparency standards: 
\begin{enumerate*}
    \item Many building energy studies face legitimate constraints on data/code sharing (e.g., proprietary systems)
    \item Manuscript transparency remains fundamental even when artifacts are available.
\end{enumerate*} 
Hence, we formulate three primary research questions 
outlined below:
\begin{tcolorbox}[
    colback=lightgray,
    colframe=gray!40,
    title=\textbf{Primary Research Questions},
    fonttitle=\bfseries,
    coltitle=black,
    boxrule=0.3mm,
    arc=2mm,
    left=2mm,
    right=2mm,
    top=1mm,
    bottom=1mm,
    enhanced
]
\RQOne\\

\RQTwo\\

\RQThree

\end{tcolorbox}
We queried three major research databases, IEEE Xplore, ACM Digital Library, and Scopus, to retrieve peer-reviewed conference articles published between 2014 and 2024. We specifically focused on studies addressing FDD in HVAC systems. The initial phase involved a broad retrieval of articles, followed by the categorization of FDD methodologies according to their type. From an initial pool of 366 unique articles, we applied defined inclusion criteria, resulting in 107 eligible articles (32\% of the original set) for the meta-review. Subsequently, ML-based studies or studies that include at least one ML technique in the proposed FDD approach were identified as the primary studies, in total 65 studies, which were then reviewed in detail with a focus on reproducibility analysis.
The detailed review process, reproducibility dimensions, and the resulting findings are presented in subsequent sections of this article. To the best of our knowledge, this is the first empirical study in the field to conduct a reproducibility analysis of machine learning FDD methods for HVAC systems.

In the following, Section~\ref{sec:related-work} reviews related work on reproducibility and existing research on FDD in building systems. Section~\ref{sec:foundations-reproducibility-dimensions} introduces the reproducibility characteristics and their relevance to this study, followed by a detailed description of the research methodology in Section~\ref{sec:research-methodology}. The findings and results from our empirical analysis are presented in Section~\ref{sec:results-and-findings}. We then share insights and recommendations for researchers in Section~\ref{sec:discussion}, and conclude this work in Section~\ref{sec:conclusion}.

\section{Background \& Related Work}
\label{sec:related-work}

\subsection{Reproducibility: A growing concern in AI and Engineering Research}
\label{sec:reproducibility-ai-eng-research}
Reproducibility grants credibility, rigor, and confidence to findings published in scientific community. It serves as a cornerstone of scientific research by ensuring transparency and enabling validation claims made by the authors. Unfortunately, this fundamental aspect is often treated as an afterthought, despite its critical role in enabling the discovery of new scientific phenomena and the ground breaking technologies through the accumulation and extension of prior findings. However, over the past two decades, it has become widely acknowledged among scientists that many studies are difficult, or in some cases nearly impossible to reproduce accurately. 

Back in 2016, over 1,500 scientists participated in a survey~\citep{baker20161}, and the responses revealed concerning insights: 70\% reported being unable to reproduce experiments published by other authors, and more than half admitted to failed attempts at reproducing their own experiments. Gundersen and Kjensmo~\citet{gundersen2018state} reviewed 400 studies published in highly ranked international artificial intelligence (AI) conferences (IJCAI\footnote{\href{https://www.ijcai.org}{International Joint Conference on Artificial Intelligence}} \& AAAI\footnote{\href{https://aaai.org/}{Association for the Advancement of Artificial Intelligence}}). Their analysis revealed that none of the reviewed studies fully shared the details required to reproduce the experiments. More specifically, only 20\% to 30\% of the necessary information was typically shared. In another study~\citep{raff2019step}, the authors aimed to quantify reproducibility by manually attempting to implement 255 studies published between 1984 and 2017. They coined the term~\emph{independent reproducibility}, referring to reproducing results without using the code provided by the original authors, as releasing code is sometimes insufficient~\citep{drummond2009replicability}. Their findings indicated a reproducibility rate of approximately 63\%, which is significantly higher than a previous study published by Gundersen~\citep{gundersen2018state}. Their significance testing further revealed that the strongest empirical relationship was associated with the~\emph{readability} of the papers. They suggested focusing on clear and detailed communication of implementation details.~We adopt a similar principle in this study, emphasizing that properly written, well-documented, and easy-to-follow papers should be a priority when publishing articles.~\citet{pham2020problems} conducted an extensive experimental study comprising 2,304 identical runs (144 experimental sets with 16 runs each), requiring over 6.5 months of GPU time. Well-known datasets~\footnote{MINST, CIFAR-10, and CIFAR100} were evaluated using six popular deep learning models~\footnote{LeNet-1, LeNet-4, LeNet-5, ResNet-38, ResNet-56, and WideResNet-28-10} using three widely used deep learning libraries~\footnote{Tensorflow, CNTK, and Theano}. The analysis revealed that, even under default identical training conditions without controlling for NI-factors~\footnote{Non determinism-introducing factors, such as shuffling, weight initialization, data augmentation, etc., affecting the training and final model accuracy.}, the accuracy gap between the least and most accurate models could reach as high as 10.8\%, even after excluding weak models, i.e., those achieving accuracy below 20\%. This striking result highlights the challenges inherent in deep learning research, challenges that, unfortunately, are often overlooked during the development and subsequently during the publishing process. 

The share of reproducibility studies is comparatively higher in other scientific disciplines than in energy related research. Researchers in the energy sector appear to be progressively acknowledging the challenges of reproducibility. However, to the best of our knowledge, no empirical studies, especially including machine learning techniques, to date have conducted an in-depth analysis of published articles in this domain to assess the state of reproducibility. That said, we did identify a few studies that offer guidelines aimed at addressing reproducibility challenges in energy research by promoting greater openness and methodological robustness~\citep{huebner2021improving, henry2021promoting, verticchio2024current, shekhorkina2024scalability}.

\citet{huebner2021improving} outlines several challenges to reproducibility and proposed adoption of the~\emph{TReQ} approach-Transparency, Reproducibility, and Quality. They identified key challenges, including domain contextual sensitivity, the high cost and time demands of trials, and resistance from external partners (e.g., utility companies) who may oppose data sharing due to competitive concerns or non-disclouse agreements (NDAs). However, the proposed solutions through the adoption of~\emph{TReQ} approach. Specifically, they recommend the preregistration of studies and their plans (PAPs), adherence to reporting guidelines, i.e., standardized indicators specifying which details should be included in published reports and early submission of research findings for community scrutiny through preprints. 

The work by~\citet{henry2021promoting} highlights several challenges inherent in energy systems modeling and model intercomparison efforts within electric sector. For instance, many energy systems have parametric and structural complexity, which makes model-based approaches more challenging compared to model-free (machine learning). As a results model designed to answer similar questions often result in dissimilar outcomes due to diverging input parameters and structural uncertainty. They presented a benchmarking framework using simplified scenarios applied to four open-source models of the U.S. electric sector. Their finding demonstrate that consistency can be improved by identifying specific structural differences and reducing parametric uncertainty. The authors also highlighted, among other issues-such unreported uncertainties and non-unique solutions in optimization problems-the lack of community-wide benchmarking standard, which is critical for reproducibility. To address these challenges, they call on researchers to increase transparency, enabling verification and improving the identification of sources of discrepancy. 

More recently,~\citet{verticchio2024current} reviewed 105 articles reporting case studies on thermal comfort improvements and conservation. In total, 112 case studies were identified, as some articles included multiple studies. The primary articles reviewed spanned from 2011 to 2022 indexed in Scopus, although the inclusion criteria target publications from 1997 to 2022. The findings revealed that reproducibility is severely effected by the lack of uniform details regarding model choices, making it difficult for other researchers to replicate the exact steps or assumptions made in a given study. Furthermore, the authors highlighted the need for ``tailored comparative tests to verify the differences among simulation software tools and algorithms", in other words, standardized test suites to understand why different tools may produce dissimilar results when applied to similar problems. Finally, they call for the establishment of FAIR (findable, accessible, interoperable, and reusable) data repositories to build a critical mass of accessible and high-quality information space. 

\subsection{FDD Methods in Energy Systems}
\label{sec:fdd-energy-systems}

In the energy building systems sector, significant energy consumption and thus operational costs could be reduced by optimizing building control and improving the detection and mitigation of faults~\citep{melgaard_fault_2022}. The collective term for processes that (automatically) identify abnormal system behavior and determine the affected component or type of fault is Fault Detection and Diagnosis (FDD). 

One approach to FDD are rule-based methods, where thresholds or if-then logic are defined to detect irregularities. These kind of methods require extensive knowledge about the systems from a domain expert. The approach is limited in the amount of faults covered, is sensitive for parameter tuning, and scales poor to different applications/settings due to the encapsulation of system specifics. For example~\citet{SCHEIN20061485} describes a rule-based system with a pre-defined set of 28 rules and 5 operation modes to detect several types of faults. 

In model-based approaches, the physical characteristics of the system are modeled using first-principles or gray-box methods by estimating system parameters~\citep{ISERMANN200571}. However, developing a physical model can be complex and computationally expensive. One way to address the challenge is to use knowledge-based approaches, where with the help of domain experts the number of in- and outputs as well as the overall model complexity is reduced wherever possible. Nevertheless, as soon as a model is tailored to a specific application by experts, it loses scalability to other applications~\citep{Yang2014}.

In recent years, the amount of available operational data has increased significantly, leading to a growing interest in machine learning (ML) approaches. These approaches aim to learn and predict the system's characteristics and its behavior from historical operational data rather than relying on complex physical modeling. This also improves scalability between different buildings or systems since no expert knowledge of the systems is required. Within ML-based FDD approaches, a variety of different methods (or a combinations of them) can be used. Prominent examples include clustering algorithms, principal component analysis (PCA), support vector machines (SVM), and neural networks (NN)~\citep{chen2023review}. However, not only the applied method matters, but also how its parameters are optimized and within what ranges. Furthermore, operational data needs to be preprocessed before model training. This preprocessing may include outlier detection, scaling or transformation, and feature selection. Since each of these steps can be implemented in various ways and with different precision, thorough documentation is essential from a reproducibility perspective. 

\begin{figure}[htbp]
    \centering
    \includegraphics[clip, trim=1.1cm 1.5cm 0.3cm 0.3cm, width=1\textwidth]{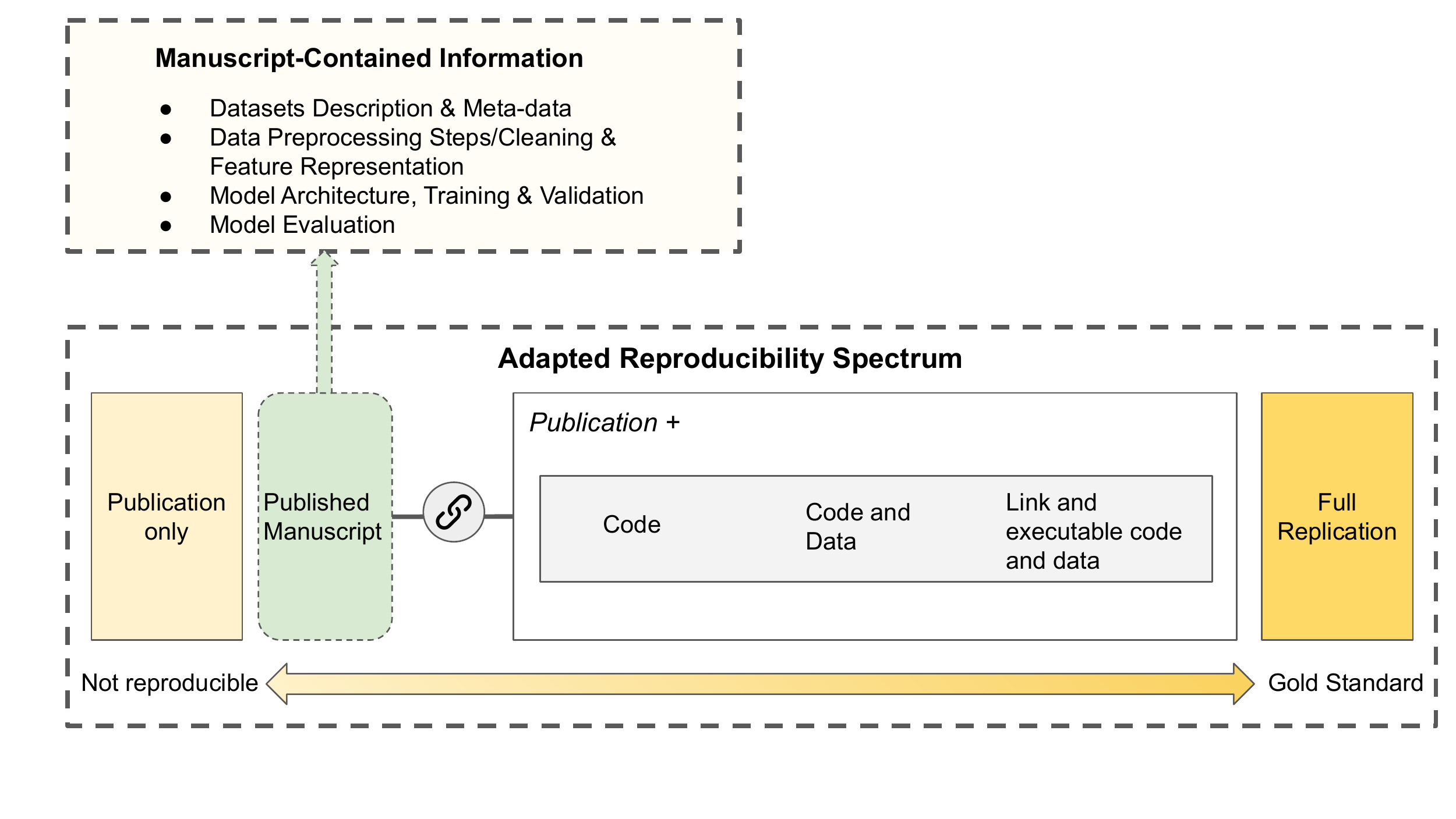}
    \caption{Adapted reproducibility spectrum from ~\citep{peng2011reproducibility}}
    \label{fig:reprod-spectrum}
\end{figure}

Currently, the application of FDD methods in industry remains limited. Most applied models still rely on expert knowledge and manually defined thresholds. However, this dependence is expected to decrease with the use of ML methods. According to~\citep{heimar_andersen_barriers_2024} key barriers for adoption include the lack of standardization of knowledge and tools, as well as, the shortage of accessible datasets and metadata.

\section{Foundations and Reproducibility Dimensions}
\label{sec:foundations-reproducibility-dimensions}

Reproducibility can rarely be assessed as a binary outcome through a simple objective function—for instance, by asking:~\textit{"Is this study reproducible?"} Unfortunately, there is no straightforward answer to such simplistic question. As it largely depends on the availability and quality of relevant artifacts and documentation presented in the article. The content within the manuscript alone is insufficient unless it is complemented by external resources, such as shared datasets, code repositories, and related materials. Crucially, the sharing of these resources does not, by itself, ensure reproducibility due to the stochastic nature of machine learning methodologies such as data shuffling, weight initialization~\citep{raste2022quantifying}, subtle differences in package versions, data leakage~\citep{semmelrock2025reproducibility} and more. Therefore, we believe, to address these nuances with fairness and transparency, an empirical study is required to quantitatively assess the \emph{degree} of reproducibility of each study based on a set of reproducibility \emph{dimensions} \citep{gundersen2018state} and a corresponding checklist (as we call them \emph{variables}). Thus, in this study, we build upon the concept of the reproducibility spectrum (\emph{degree}) presented by~\citep{peng2011reproducibility}, as shown in Figure~\ref{fig:reprod-spectrum}. Our goal is to identify and evaluate the information provided within the manuscript that is critical in supporting, and to some extent ensuring, the reproducibility of the study. In particular, we focus on identifying information related to datasets, data pre-processing steps, model development, hyperparameters optimization, evaluation strategies, and accessibility of external artifacts such as executable code, datasets, pre-processing and fine-tuning implementations.

To answer primary research questions, we curated a reproducibility variables checklist adapted from existing reproducibility surveys and guidelines~\citep{gundersen2018state, artrith2021best, pineau2021improving, mukhtar2024reproducibility, olszewski2023get}, specifically for the task of machine learning-based FDD in HVAC energy systems. First, we present a general categorization of the variables into several classes, including dataset, data cleaning/pre-processing, model development, evaluation, and code availability. These variables along with their description and categories are presented in Table~\ref{tab:reprod-vars}. The descriptions of these variables are framed as questions to guide the review process, with answers selected based on predefined criteria. For clarity and ease of reference during the discussion of results, we assign each variable a unique identifier. We then group these variables into three reproducibility dimensions as presented by~\citep{gundersen2018state}: \begin{enumerate*}
    \item data
    \item methodology
    \item experiment.
\end{enumerate*}

It is important to note that we make a distinction between the categories of variables (Table~\ref{tab:reprod-vars}) and the reproducibility dimensions. The categories represent different stages in the development workflow of ML-based methodologies.~In contrast, the dimensions are an overlapping subset drawn from across categories. These dimensions highlight the critical aspects that jointly influence the reproducibility of a study. As discussed previously, we follow the approach of~\citep{gundersen2018state} in categorizing reproducibility into three distinct dimensions. We assess each dimension using the concept of reproducibility degree~\citep{peng2011reproducibility}, where a higher degree indicates that the reviewed articles offer more comprehensive information and materials to support reproducibility. While preserving the core intent of these dimensions, we carefully reinterpret and refine their definitions to better align with our study's scope. Since our approach involves thorough manual article review, \emph{without conducting direct experimental replication}, we assess reproducibility variables through this evaluation. Recall that the guiding principle of this study is that all necessary concrete information and materials for reproducibility should be transparently documented, clearly structured, and straightforward to understand. Furthermore, across all dimensions, the studies' objective, i.e., ML based FDD in HVAC systems, serves as a foundational requirement. We now define the reproducibility dimensions as follows: 

\vspace{1em}
\DimensionOne

\vspace{1em}

\DimensionTwo

\vspace{1em}

\DimensionThree


\newcommand{\dataListed}{\ensuremath{data_{listed}}}
\newcommand{\dataMetadata}{\ensuremath{data_{metadata}}}
\newcommand{\dataStats}{\ensuremath{data_{stats}}}
\newcommand{\dataType}{\ensuremath{data_{type}}}
\newcommand{\dataAccess}{\ensuremath{data_{access}}}

\newcommand{\preprocessData}{\ensuremath{preproc_{data}}}
\newcommand{\preprocessFeatures}{\ensuremath{preproc_{features}}}
\newcommand{\multipleData}{\ensuremath{multiple\ data}}

\newcommand{\optMentioned}{\ensuremath{opt_{mentioned}}}
\newcommand{\optBaseline}{\ensuremath{opt_{baseline}}}
\newcommand{\optProcedure}{\ensuremath{opt_{procedure}}}
\newcommand{\paramsModels}{\ensuremath{params_{models}}}
\newcommand{\paramsBaseline}{\ensuremath{params_{baseline}}}
\newcommand{\paramsBestmodel}{\ensuremath{params_{best\ model}}}
\newcommand{\paramsBestbaseline}{\ensuremath{params_{best\ baseline}}}

\newcommand{\evalSplitting}{\ensuremath{eval_{splitting}}}
\newcommand{\evalMetrics}{\ensuremath{eval_{metrics}}}
\newcommand{\evalSigtest}{\ensuremath{eval_{sig\ test}}}

\newcommand{\codeLink}{\ensuremath{code_{link}}}
\newcommand{\codeEmpty}{\ensuremath{code_{empty}}}
\newcommand{\codePreprocess}{\ensuremath{code_{preproc}}}
\newcommand{\codeFeaturegen}{\ensuremath{code_{feature\ gen}}}
\newcommand{\codeEval}{\ensuremath{code_{eval}}}
\newcommand{\codeParamasopt}{\ensuremath{code_{params\ opt}}}
\newcommand{\codeInfo}{\ensuremath{code_{info}}}
\newcommand{\codeRunnable}{\ensuremath{code_{runnable}}}

\begin{longtable}{p{2.5cm}|p{10cm}|C{4cm}}
\caption{Categorization of Reproducibility Variables}
\label{tab:reprod-vars} \\
\toprule
\rowcolor[HTML]{C0C0C0} 
\multicolumn{3}{c}{\cellcolor[HTML]{C0C0C0}\textbf{Dataset}} \\ \midrule
\rowcolor[HTML]{EFEFEF} 
\multicolumn{1}{c|}{\cellcolor[HTML]{EFEFEF}\textbf{Identifier}} & \multicolumn{1}{c|}{\cellcolor[HTML]{EFEFEF}\textbf{Variables}} & \multicolumn{1}{c}{\cellcolor[HTML]{EFEFEF}\textbf{Answers}} \\ 
\midrule

\dataListed & Is the dataset (or datasets) listed? & 1 / 0  \\ 
\dataMetadata & Are the metadata and description of the dataset (or datasets) provided? & 1 / 0  \\
\dataStats & Are the relevant statistics discussed, e.g., the number of samples, etc.? & 1 / 0 \\
\dataType & What is the type of the dataset(s)? & Real-world / Simulation / Experiment / No Information \\
\dataAccess & Is information about the accessibility of the dataset(s) shared? & Purchasable / Public / Proprietary / No Information \\
\midrule

\rowcolor[HTML]{C0C0C0} 
\multicolumn{3}{c}{\cellcolor[HTML]{C0C0C0}\textbf{Data Cleaning/Preprocessing \& Feature Representations}}\\ \midrule
\rowcolor[HTML]{EFEFEF} 
\multicolumn{1}{c|}{\cellcolor[HTML]{EFEFEF}\textbf{Identifier}} & \multicolumn{1}{c|}{\cellcolor[HTML]{EFEFEF}\textbf{Variables}} & \multicolumn{1}{c}{\cellcolor[HTML]{EFEFEF}\textbf{Answers}} \\ 
\midrule
\preprocessData & Are the data preprocessing steps documented? & 1 / 0 \\
\preprocessFeatures & Are data to feature representation methods clearly described? & 1 / 0 \\
\multipleData & If multiple data sources are used, is their integration clearly stated? & 1 / 0 / No Information \\
\midrule

\rowcolor[HTML]{C0C0C0} 
\multicolumn{3}{c}{\cellcolor[HTML]{C0C0C0}\textbf{Model Training \& Validation}}\\ \midrule
\rowcolor[HTML]{EFEFEF} 
\multicolumn{1}{c|}{\cellcolor[HTML]{EFEFEF}\textbf{Identifier}} & \multicolumn{1}{c|}{\cellcolor[HTML]{EFEFEF}\textbf{Variables}} & \multicolumn{1}{c}{\cellcolor[HTML]{EFEFEF}\textbf{Answers}} \\ 
\midrule
\optMentioned & Is hyperparameter optimization mentioned for the proposed model(s)? & 1 / 0 \\
\optBaseline & Is hyperparameter optimization mentioned for the baselines? & 1 / 0 \\
\optProcedure & Is the hyperparameter optimization procedure described? & 1 / 0 \\
\paramsModels & Are hyperparameter search ranges reported for the proposed model? & 1 / 0 \\
\paramsBaseline & Are hyperparameter search ranges reported for the baselines? & 1 / 0 \\
\paramsBestmodel & Are the best hyperparameters reported for the proposed model? & 1 / 0 \\
\paramsBestbaseline & Are the best hyperparameters reported for the baselines? & 1 / 0 \\

\rowcolor[HTML]{C0C0C0} 
\multicolumn{3}{c}{\cellcolor[HTML]{C0C0C0}\textbf{Evaluation}}\\ \midrule
\rowcolor[HTML]{EFEFEF} 
\multicolumn{1}{c|}{\cellcolor[HTML]{EFEFEF}\textbf{Identifier}} & \multicolumn{1}{c|}{\cellcolor[HTML]{EFEFEF}\textbf{Variables}} & \multicolumn{1}{c}{\cellcolor[HTML]{EFEFEF}\textbf{Answers}} \\ 
\midrule

\evalSplitting & What type of data splitting is reported? & Single split / Train-Test-Validation / Cross Validation / No Information\\
\evalMetrics & Are the metrics used for evaluation reported? & 1 / 0 \\
\evalSigtest & Are the details of statistical significance testing provided? & 1 / 0 \\

\rowcolor[HTML]{C0C0C0} 
\multicolumn{3}{c}{\cellcolor[HTML]{C0C0C0}\textbf{Code Repository}}\\ \midrule
\rowcolor[HTML]{EFEFEF} 
\multicolumn{1}{c|}{\cellcolor[HTML]{EFEFEF}\textbf{Identifier}} & \multicolumn{1}{c|}{\cellcolor[HTML]{EFEFEF}\textbf{Variables}} & \multicolumn{1}{c}{\cellcolor[HTML]{EFEFEF}\textbf{Answers}} \\ 
\midrule

\codeLink & Is the link to the code repository available? & 1 / 0 \\
\codeEmpty & Is the code repository empty? & 1 / 0 \\
\codePreprocess & Is the source code for data pre-processing provided in the repository? & 1 / 0 \\
\codeFeaturegen & Is the source code to generate features of the dataset provided? & 1 / 0 \\
\codeEval & Is the source code for evaluation provided in the repository? & 1 / 0 \\
\codeParamasopt & Is the source code for hyperparameter tuning provided in the repository? & 1 / 0 \\
\codeInfo & Is supplementary code info (e.g., README, requirements.txt) provided? & 1 / 0 \\
\codeRunnable & Is a runnable model implementation provided (e.g., Docker)? & 1 / 0 \\
\bottomrule
\end{longtable}

\begin{figure}[htbp]
    \centering
    \includegraphics[clip, trim=0cm 0cm 0cm 0cm, width=1\textwidth]{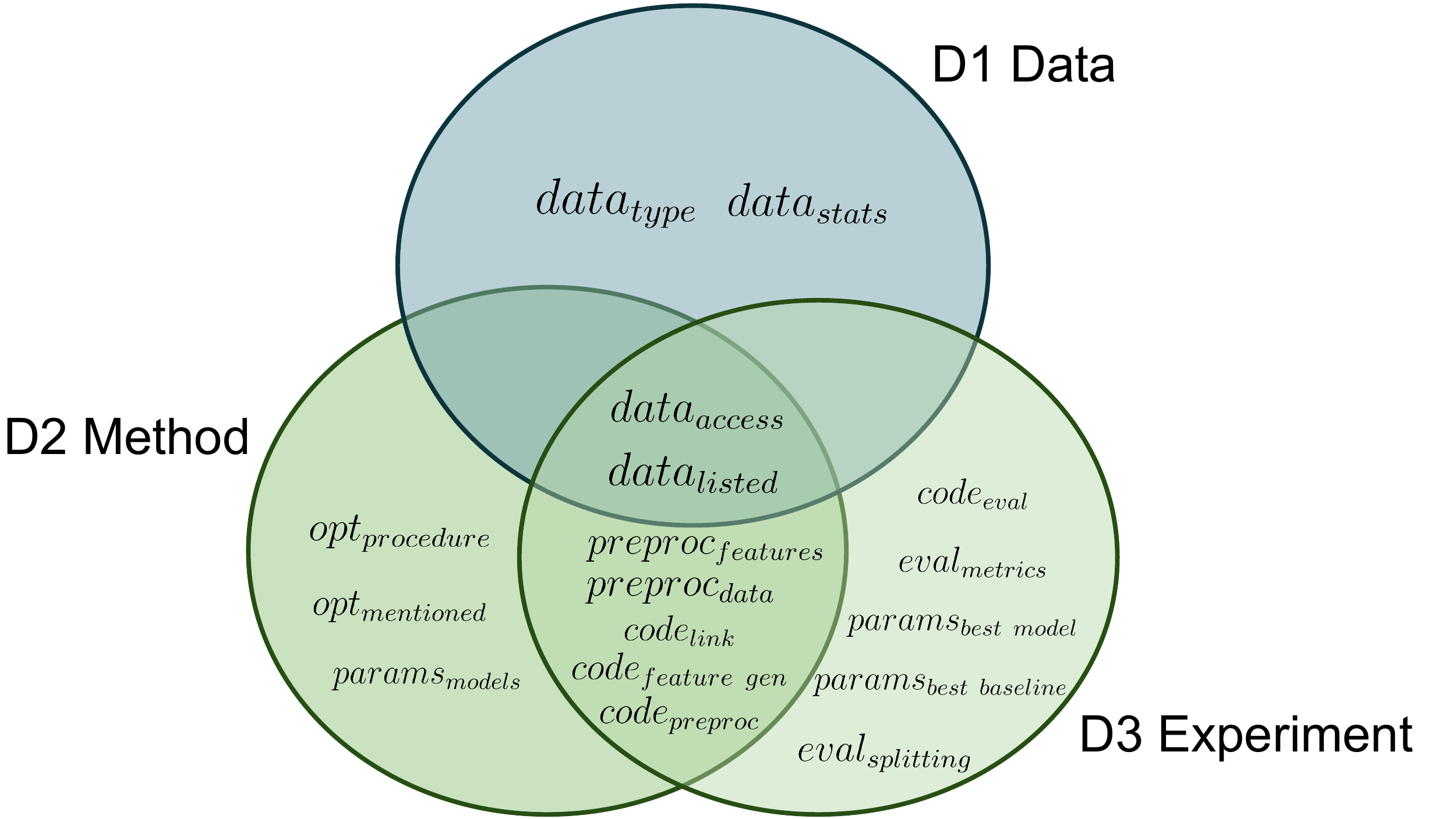}
    \caption{Reproducibility Dimensions}
    \label{fig:reprod-dimensions}
\end{figure}

These dimensions collectively determine the overall degree of reproducibility. For example, a study that provided details on hyperparameter optimization and includes the corresponding code package may still be irreproducible if no information is provided for the dataset used. 
These dimensions are overlapping subsets derived from the reproducibility variables as shown in Figure~\ref{fig:reprod-dimensions}. The figure presents the list of \emph{variables} and illustrates their association with the three dimensions, including overlaps across them. Dimensions $D_{2}$ and $D_{3}$, referring to the method and experiment, respectively, are predicated on the availability of the dataset(s), i.e., $D_{1}$. The methodologies proposed in scientific articles, especially those based on machine learning, are often presented as frameworks, pipelines, or algorithmic pseudocode. Accordingly, we identified key variables such as data processing, feature generation, hyperparameter optimization, ranges for hyperparameters during the optimization, and the availability of a code repository link, all of which are essential for reproducing the method. This dimension also intersects with Dimension $D_{1}$, as re-running the methodology requires access to the original datasets. As for dimension $D_{3}$, we consider it the desirable outcome of reproducibility, as it focuses on replicating the results of the proposed technique, comparing them with established baselines, and validating the experimental outcomes reported in the article. This dimension includes variables related to dataset accessibility, feature generation, data preprocessing, the corresponding coding scripts, evaluation strategies, data splitting techniques, and the best-performing parameters for both the proposed and baseline models.

So far, we have described the reproducibility \emph{variables} (Table~\ref{tab:reprod-vars}) and defined the dimensions that span these variables. The rationale behind dissecting the reproducibility variables into three overlapping subsets is to assess the degree of reproducibility, i.e., how well the variables within each dimension are documented. Our next objective is to quantify each proposed study by assigning a reproducibility score, reflecting the extent to which information and materials are shared to support reproducibility. To achieve this, we adapt the quantification scheme from~\citep{gundersen2018state}. We first evaluate each dimension separately, followed by computing an overall aggregated reproducibility score across all dimensions. For a given study $s$, we first define a function $v_j(s)$ for the $j^{th}$ variable as follows: \[
v_j(s) =
\begin{cases}
1, & \text{if study } s \text{ reports variable } j \\
0, & \text{otherwise OR ``No Information"} \\ 
\end{cases}
\]

The dimensions $D_{1}$ (Data), $D_{2}$ (Method), and $D_{3}$ (Experiment) are quantified as follows: for a given dimension $i$ where $i \in \{1,2,3\}$

\begin{tcolorbox}[colback=grey!2, colframe=black, boxrule=0.5pt]
\begin{equation}
D_i(s) = \frac{\sum_{j=1}^{D_{i}} v_j(s)}{|D_{i}|}
\label{equ:primary-equation}
\end{equation}
\end{tcolorbox}

For clarity and brevity, a study $s$ is assigned a score of 1 for a dimension only if all variables within that dimension are documented. In other words, it represents the proportion of variables documented for dimension $i$, ranging from 0 (no documentation) to 1 (full documentation).
However, to quantify how many studies fully document all variables across dimensions, i.e., reproducible, we define the degree as follows:

\begin{equation}
    degree(s) = \frac{\sum_{j=1}^{V} v_{j}(s)}{|V|}
    \label{equ:degree-calc}
\end{equation}
where $V = D_{1}\ \cup D_{2}\ \cup D_{3}$.
In our study, just like in~\citet{gundersen2018state}, we treat each variable equally and assign uniform weight, i.e., $1$.~The reason for treating them as equally important is that the variables are carefully curated to reflect fundamental aspects of research methodology in the building system scientific discipline for the ML-based FDD. For example, one of the key barriers to open science in building system field is the use of proprietary datasets, or datasets that cannot be made openly available due to non-disclosure agreements (NDAs) and strict data-sharing policies and regulations. To address this, we assign a score of 0 only when no information about such details is provided. However, in cases where the dataset is, for example, purchasable, we still consider this crucial information and record 1—provided that it is explicitly mentioned in the study and includes a reference or external link. Such documentation enables independent research teams to access the dataset if needed. 

We conclude this section by emphasizing that information in a manuscript should be clearly reported and easy to find. Clear documentation is essential for ensuring the reproducibility of proposed studies, while ambiguity challenges the ability to achieve it. To address this, we carefully designed a list of reproducibility variables, organized them according to key aspects of research methodology, and quantified them to provide insights into the current state of reproducibility in building system for ML-based FDD in HVAC systems. In the next section, we detail our survey methodology, including the retrieval of articles and the manual review process.
\section{Research Methodology}
\label{sec:research-methodology}
\begin{figure}[htbp]
    \centering
    \includegraphics[clip, trim=3.0cm 0.9cm 2cm 1.3cm, width=1\textwidth]{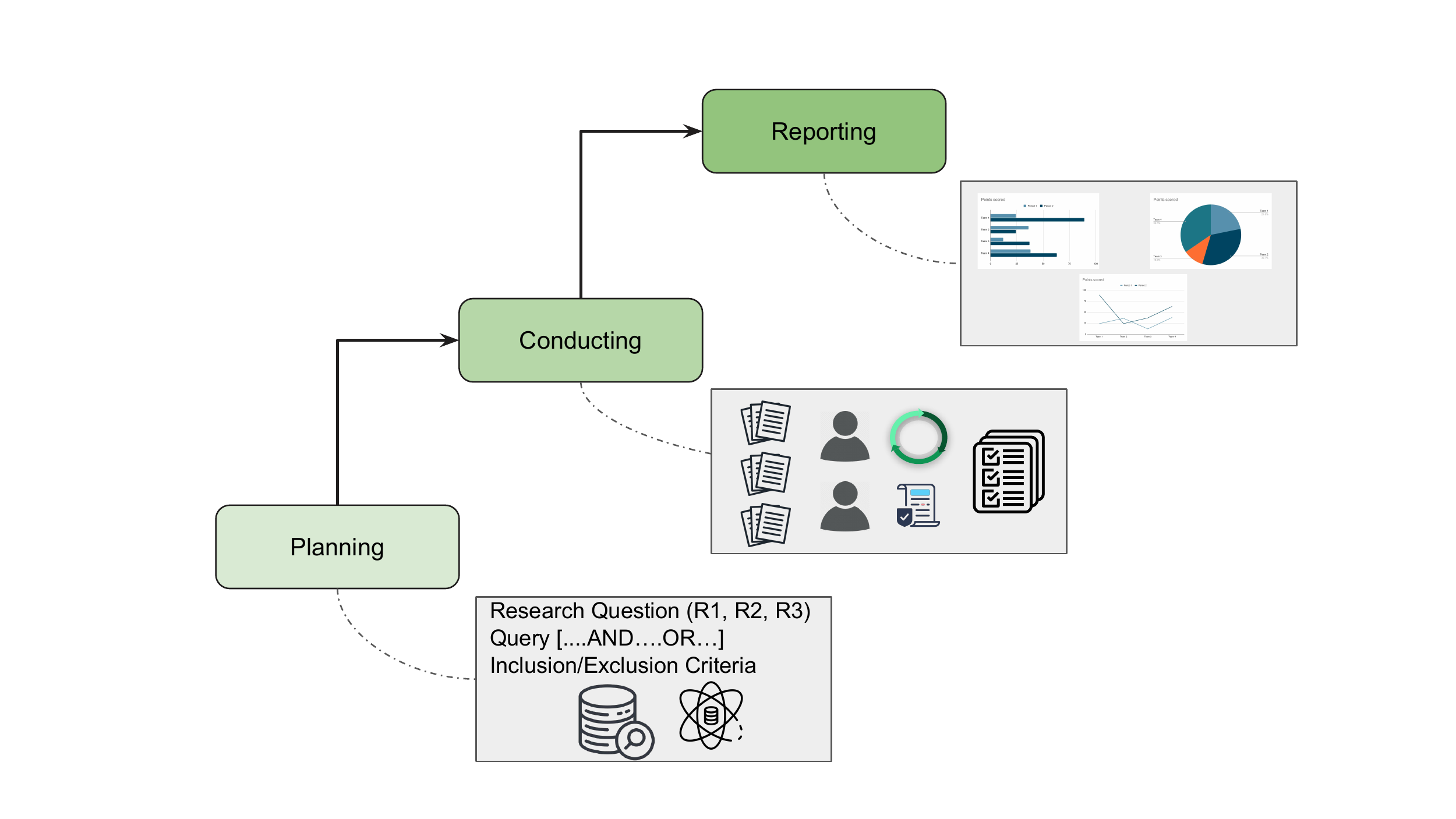}
    \caption{Research Methodology}
    \label{fig:research-methogolody}
\end{figure}

This study adopts a comprehensive literature review process~\citep{budgen2006performing} to guide the articles review process focused on the application of machine learning-based FDD for HVAC systems in building system. The review process is structured into three main phases: \begin{enumerate*}
    \item \emph{planning the review}
    \item \emph{conducting the review}, and
    \item \emph{reporting the review}.   
\end{enumerate*}
Figure~\ref{fig:research-methogolody} provides a bird's-eye view of the research methodology used in this study.

In the first phase, \emph{planning}, research questions are defined (see Section~\ref{sec:objectives-contributions}) alongside the development of a search protocol. This includes identifying relevant keywords and selecting scientific databases. Once the initial pool of articles is retrieved by querying these databases using defined inclusion and exclusion criteria, a preliminary categorization is conducted to filter articles that fit the study's objectives and research questions. A meta-analysis of this categorization is presented later in Section~\ref{subsec:results-overview-fdd}. In the second phase, \emph{conducting}, the manual review, by two researchers, is carried out in three rounds. The details of this phase are provided in Section~\ref{subsec:data-extraction-reviewing-process}. Finally, in the third phase, \emph{reporting}, the results and key insights are presented and discussed. 

\subsection{Databases \& Search Strategy}  
\label{subsec:search-protocols-databases}
To ensure comprehensive coverage, the following scientific databases are used to retrieve relevant articles in this study: IEEE Xplore, ACM Digital Library, and Scopus. The main reason for this selection is their extensive collections of scientific publications spanning a wide range of disciplines, including computer science, engineering, artificial intelligence, and other multidisciplinary fields.

Before designing the search query to retrieve articles from the selected databases we consulted with the domain experts in building system to identify and shortlist relevant keywords related to the domain. This preliminary step is taken to warrant a certain level of confidence that the retrieved results, i.e., articles, are relevant and aligned with the scope and objectives of this study. The list of identified key terms is presented in Table~\ref{tab:key-terms}. Based on this shortlisting, the final search query is constructed using a combination of logical operators, mainly conjuctions and disjunctions, and then used to query the databases. The complete query is presented below:
\begin{tcolorbox}[colback=lightgray, colframe=black, boxrule=0.5pt]

(fault detection and diagnosis OR fdd OR fault detection and diagnosis) AND \\
(hvac OR heating ventilation and air conditioning OR chillers OR ahu OR air handling unit OR vav OR variable air volume OR air conditioning) AND \\
(building energy systems OR building systems OR energy building OR buildings)
\end{tcolorbox}

\begin{table}[]
\begin{tabular}{|c|c|c|c|}
\hline
fault detection and diagnosis & fdd                                      & fault detection    & fault diagnosis  \\ \hline
hvac                          & heating ventilation and air conditioning & chillers           & ahu              \\ \hline
air handling unit             & vav                                      & variable air volume & air conditioning \\ \hline
building energy systems       & building systems                         & energy building    & buildings        \\ \hline
\end{tabular}
\caption{List of key terms identified}
\label{tab:key-terms}

\end{table}

\subsection{Articles Selection Criteria}  
\label{subsec:selection-criteria}
Querying scientific databases often results in many false positives, even when using relevant keywords, and may also return outdated or irrelevant publications. For example, in our case, we are specifically interested in reviewing articles published in conferences within a defined date range. Therefore, it is important to establish clear and well-defined selection criteria to ensure that the results remain aligned with the predetermined objectives and goals of the study. More specifically, we define our inclusion criteria such that if a paper meets these conditions then it is selected for review, consequently, its negation serves as the exclusion criteria and must be false. The inclusion criteria are as follows:
The article
\begin{enumerate}
    \item is related to energy building systems.
    \item proposes a methodology for heating ventilation and air conditioning systems.
    \item presents an application for fault detection and diagnosis in HVAC systems.
    \item is published in the conference proceedings.
    \item is published between years 2014 and 2024.
    \item is written in English.
    \item is not a survey paper.
\end{enumerate}

Once the search strategy, database selection and selection criteria are established, an initial set of articles are retrieved. The search query resulted in a total of 320 unique articles published in conferences related to building systems, engineering, artificial intelligence, computer science, and their interdisciplinary fields. These articles cover a range methodological approaches, including model-based (MB), knowledge-based (KB) and machine learning-based (ML) techniques.

\subsection{Articles Extraction \& Reviewing Process}  
\label{subsec:data-extraction-reviewing-process}
\begin{figure}[htbp]
    \centering
    \includegraphics[clip, trim=0.0cm 0.0cm 4cm 0.0cm, width=1\textwidth]{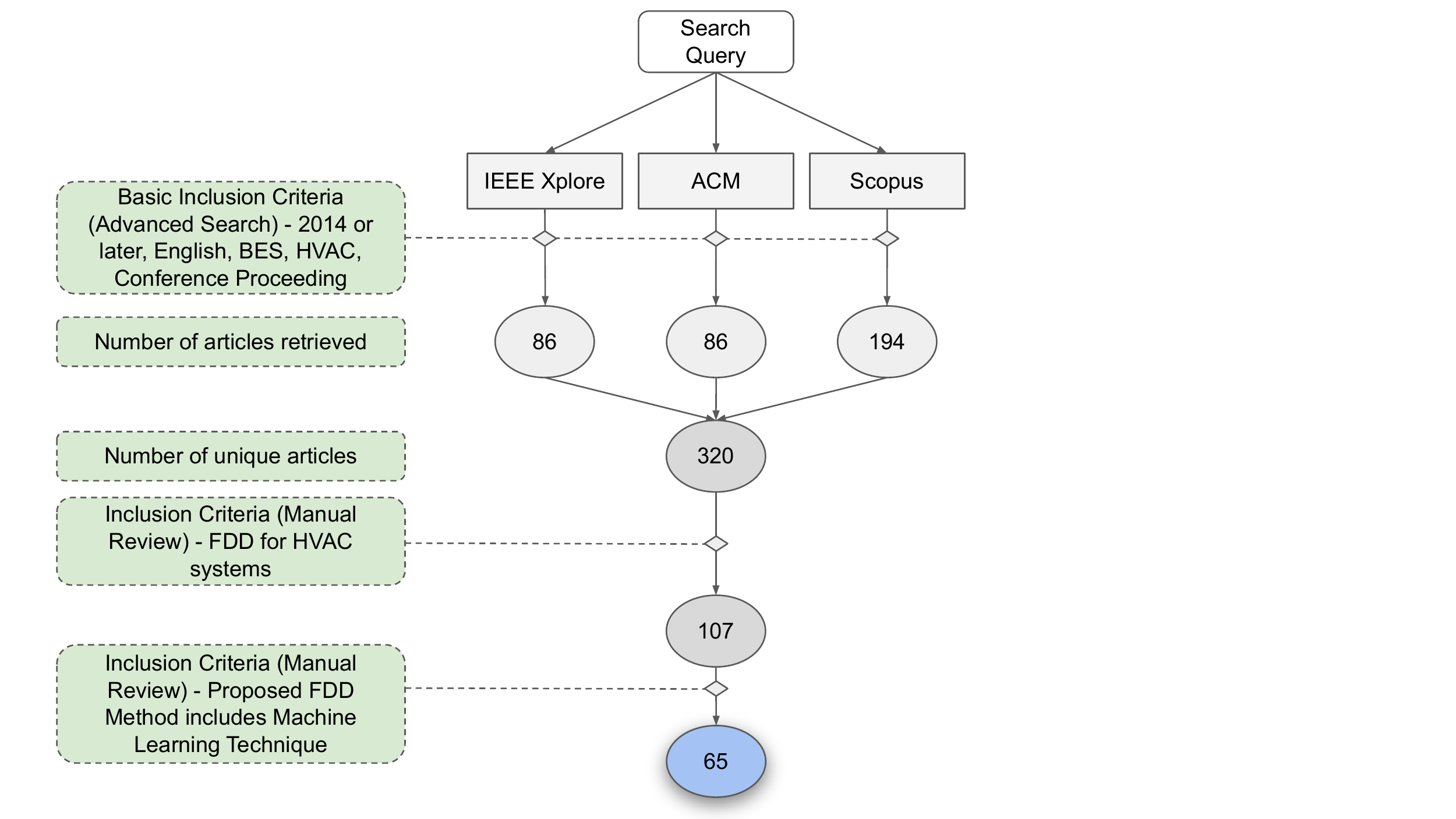}
    \caption{Articles Retrieval Process Flowchart}
    \label{fig:articles-retrieval-process}
\end{figure}

The overall steps involved in the \emph{planning} and \emph{conducting} phases, along with their outcomes, are presented in Figure~\ref{fig:articles-retrieval-process}. The query resulted in a total of 320 (out of 366) unique articles. In our experience with Scopus, similarly to~\citet{verticchio2024current}, the search results included a high number of false positives relative to the provided keywords and duplicates. This significantly increased the reviewing effort. In contrast, the articles retrieved from ACM and IEEE Xplore were generally more relevant with higher precision compared to those from Scopus. Nevertheless, the first manual review focused on classifying these articles into one of the following categories: model-based (MB), knowledge-based (KB), machine learning-based (ML), or a combination of these approaches. This classification process resulted in 107 relevant articles. From these, 65 were selected as primary articles each including at least one ML technique in the proposed methodology for fault detection and diagnosis in HVAC systems\footnote{The reviewed articles, along with the recorded reproducibility variables, are available here: \url{https://github.com/tuw-isab/reproducibility-analysis-ml-based-fdd-hvac}}.

Two authors independently reviewed the primary articles using the predefined list of reproducibility variables. As briefly mentioned earlier, this phase is completed in three rounds. In the first round, both authors reviewed a small subset of articles to establish consistent understanding of the reproducibility variables and to identify any ambiguous parts requiring clarification or refinement. In the second round, all relevant articles were reviewed independently. In the third round, the outcomes for each variable and article were compared, and any discrepancies were resolved through discussion. In general, conflicts observed during these rounds were minimal and resolved after discussions. This is mainly because of the preliminary round, which led to an improved understanding of the variables, their definitions, and the overall review process.

\section{Results \& Findings}  
\label{sec:results-and-findings}
This section begins with an overview of the methodologies employed in the past decade (Section~\ref{subsec:results-overview-fdd}). We then present our descriptive and qualitative analysis of the reproducibility variables to address the primary research questions (Section~\ref{subsec:results-reproducibility-analysis}). This is followed by the quantification of the reproducibility dimensions and an overall assessment of articles in terms of degree of sharing crucial information is presented in Section~\ref{subsec:results-assessment-reprod-dims}. 
\begin{figure}
    \centering
    \includegraphics[width=1\linewidth]{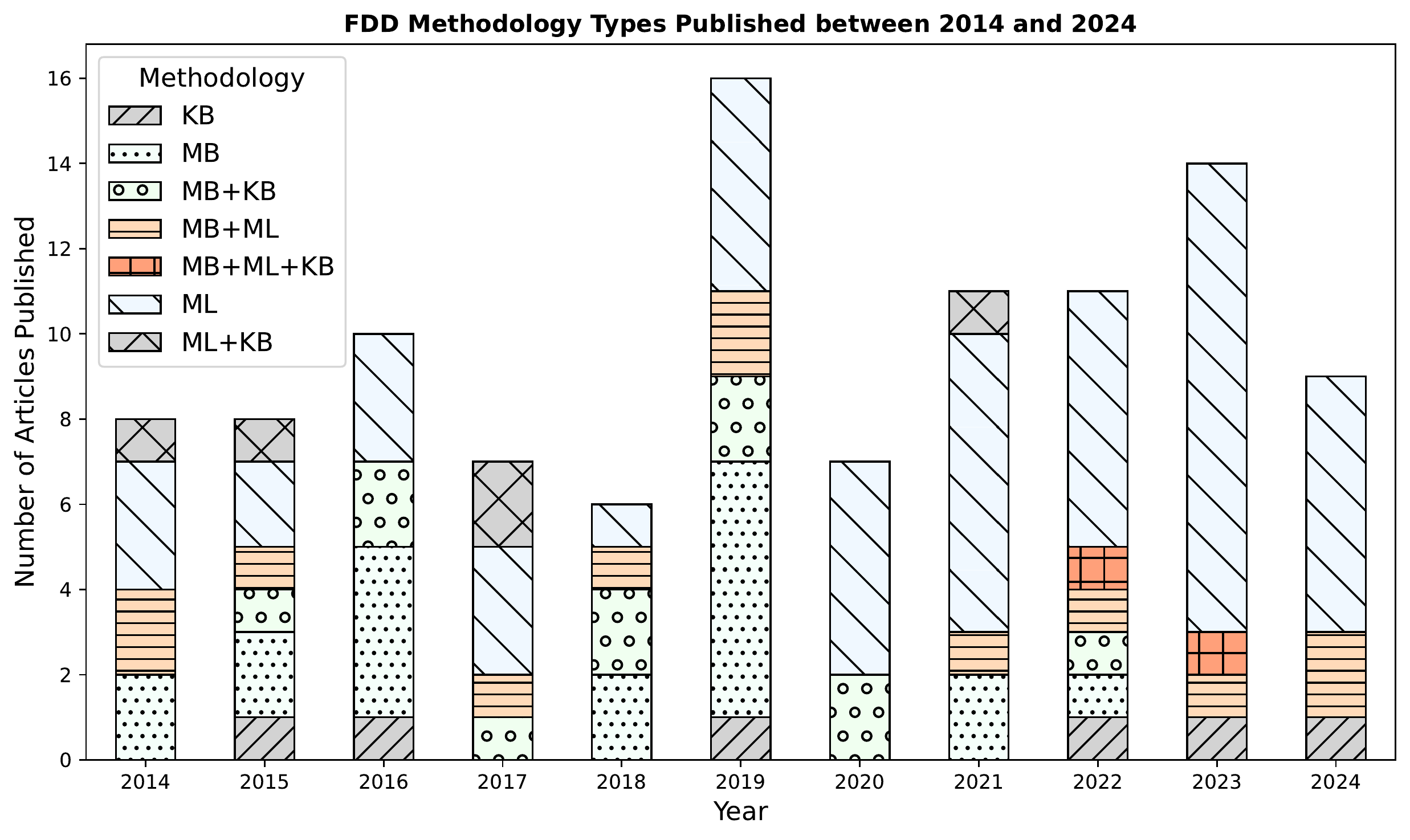}
    \caption{Number of articles and types of FDD methodologies published (2014-2024)}
    \label{fig:articles-distribution}
\end{figure}

\subsection{Overview of Fault Detection and Diagnosis Methodologies}
\label{subsec:results-overview-fdd}
Figure~\ref{fig:articles-distribution} provides an overview of the articles published between 2014 and 2024, specifically addressing FDD in HVAC systems. A key observation is the general increase in publications after the initial five years. This increase can be attributed to the growing adoption of Internet of Things (IoT) technologies, such as smart-meters, real-time monitoring, and building automation systems, which have matured significantly over the past decade~\citep{moudgil2023integration, ahmad2021using}. According to~\citep{ahmad2021using}, the number of connected IoT devices is projected to reach 75.4 billion by the end of 2025, nearly five times the number back in 2015. More importantly, numerous studies have consistently demonstrated that IoT integration in building systems contributes to substantial improvements in energy efficiency, cost reduction, and environmental sustainability; some examples are ~\citep{rohayani2024effect, poyyamozhi2024iot, moudgil2023integration}. 

A closer examination of the publication trends reveals that ML-based approaches have been consistently represented throughout the review period, exhibiting a clear upward trajectory. Standalone ML-based FDD studies account for the largest share, comprising 49\% of the publications; this share further increases up to 64\% when considering hybrid approaches that combine ML with MB and KB techniques. This growth underscores the increasing reliance on ML techniques for FDD in HVAC systems. We believe that this increase is driven by the growing recognition of the effectiveness and benefits of ML techniques across a range of research domains, including software engineering~\citep{abid2021review,aleem2015benchmarking, mukhtar2022boosting}, healthcare~\citep{maity2017machine, sharma2014brain}, and agriculture~\citep{tripathi2016recent, lu2017identification, amara2017deep}, among others. In contrast, model-based methods (18\%) and hybrid approaches (21\%), those combining model-based with either knowledge-based or machine learning components, also appear regularly in the literature but do not exhibit the same degree of growth. While these methods remain relevant, their adoption trajectory is comparatively less pronounced than that of purely ML-based techniques. Techniques that integrate all three components, MB+ML+KB, constitute only a minor portion of the literature, representing approximately 2\% of the reviewed studies. The remaining variants, including standalone KB methods and their hybrid combinations (e.g., ML+KB), collectively account for only approximately 10\% of the reviewed studies.

Publications in recent years clearly reflect a shift in methodological focus, with ML approaches increasingly surpassing model-based and knowledge-based reasoning methods in terms of research activity. In the following subsections, we turn our attention to answer the primary research questions and presenting our findings and insights related to reproducibility aspects. To this end, we systematically reviewed a subset of articles that propose FDD techniques for HVAC systems and incorporate at least one ML technique. This includes studies classified under the ML category as well as those employing hybrid approaches that combine ML with MB or KB methods.

\subsection{Reproducibility Variables Analysis}
\label{subsec:results-reproducibility-analysis}
\begin{figure}
    \centering
    \includegraphics[width=0.7\linewidth]{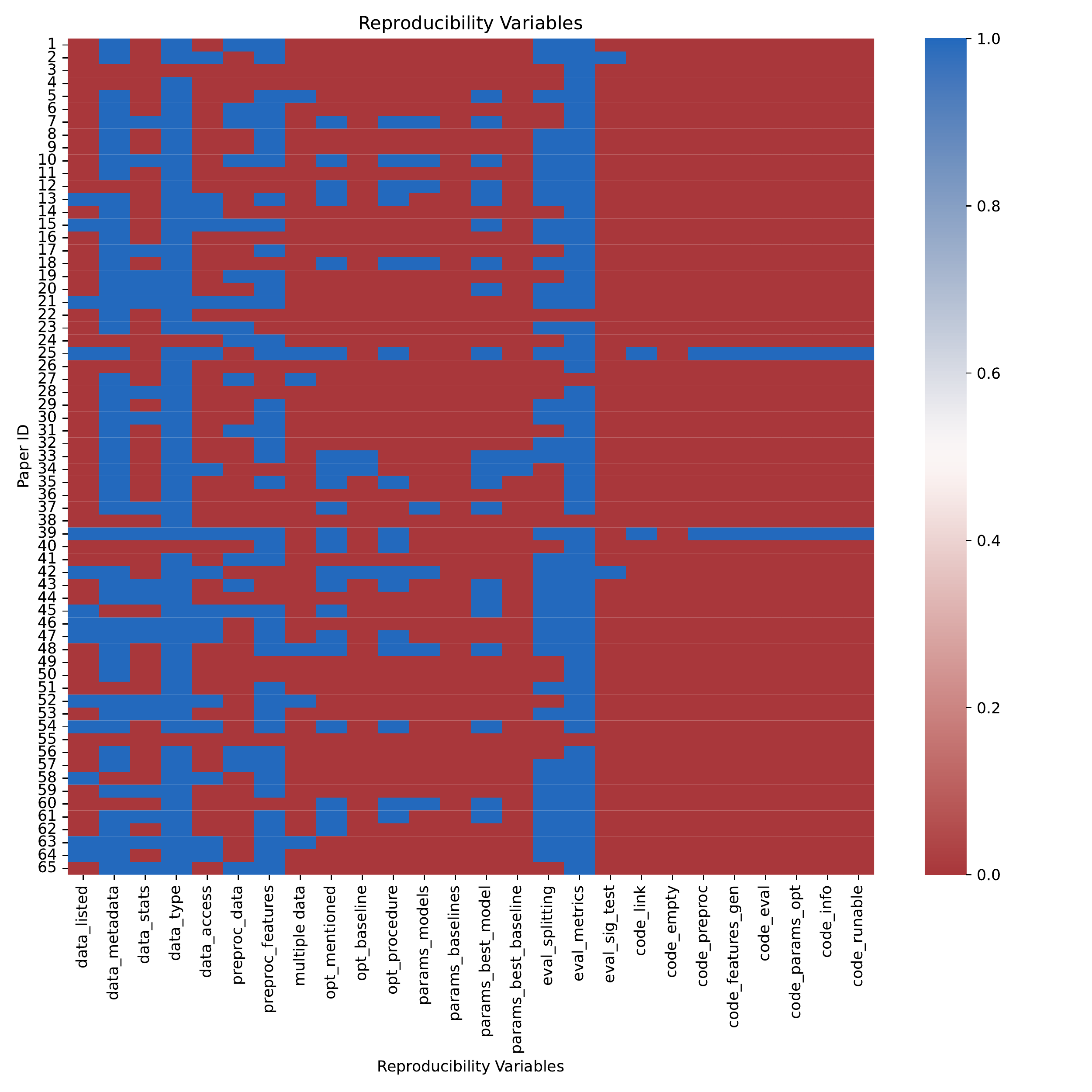}
    \caption{Heatmap of the Reproducibility Variables}
    \label{fig:heatmap-variables}
\end{figure}

We start with the information recorded for the list of variables presented in Table~\ref{tab:reprod-vars}. Figure~\ref{fig:heatmap-variables} presents, in the form of a heatmap, an overall assessment of the research articles that have documented reproducibility-related variables. It is evident that none of the variables are fully documented across the reviewed studies. In particular, information sharing related to model training and code packages appears to be among the least reported categories. This is even more pronounced in the case of code package sharing, with only a few exceptions. We now share results and analysis to answer the primary research questions.

\begin{tcolorbox}[
    colback=white,
    title=\textbf{RQ1},
    colframe=gray!40,
    fonttitle=\bfseries,
    coltitle=black,
    boxrule=0.3mm,
    arc=2mm,
    left=2mm,
    right=2mm,
    top=1mm,
    bottom=1mm,
    enhanced
]
The literature is not well-documented, and many of the variables are not fully reported across the reviewed articles. 
\end{tcolorbox}

\begin{figure}[htbp]
    \centering
    \includegraphics[clip, trim=0.0cm 0.0cm 0cm 0.0cm, width=1\textwidth]{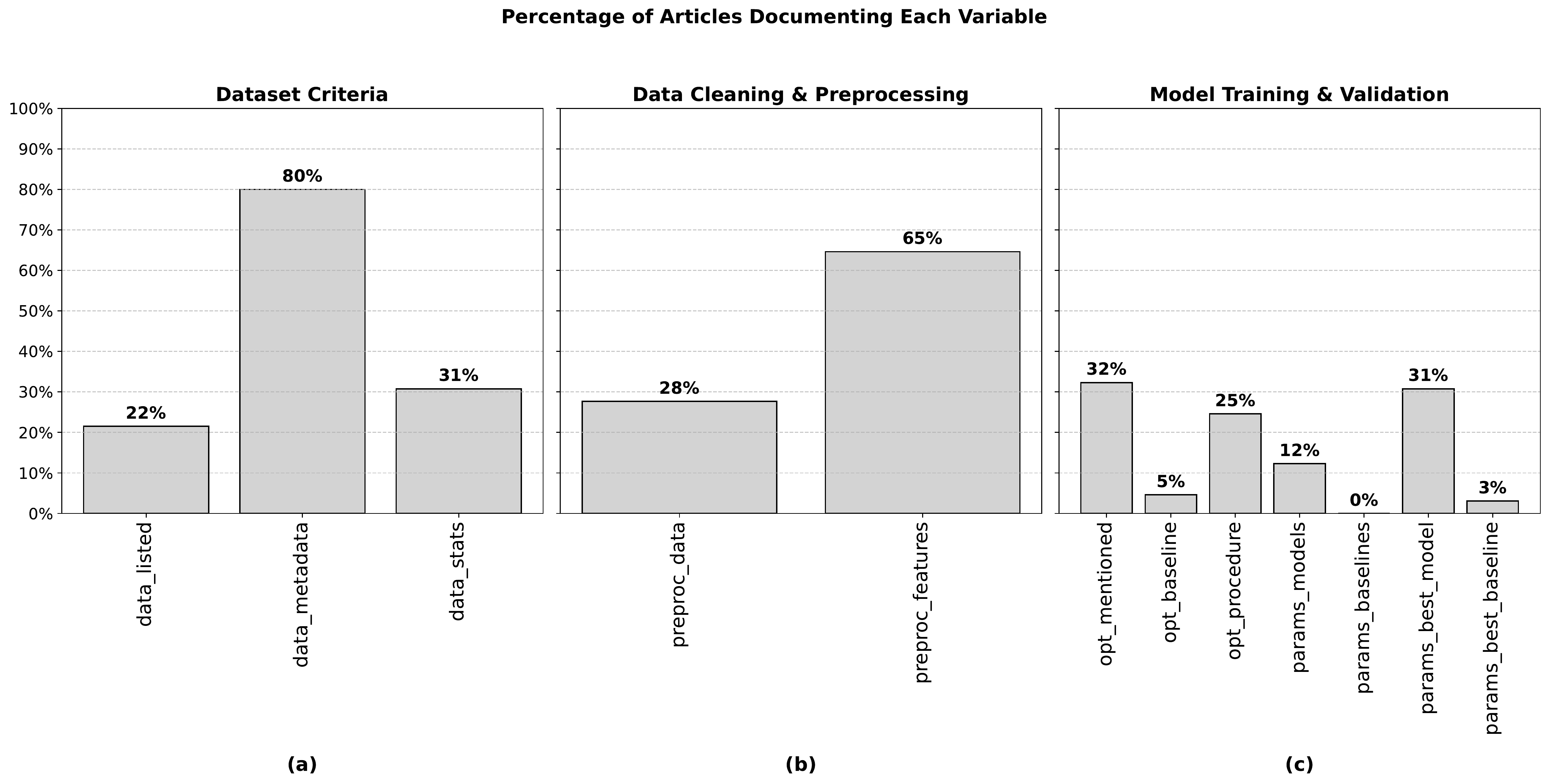}
    \caption{Percentages of Variables Recorded}
    \label{fig:variables-percentages-dataset-cleaning-training}
\end{figure}

Figure~\ref{fig:variables-percentages-dataset-cleaning-training} shows the extent to which information is documented for variables related to dataset characteristics, preprocessing, model development and its fine tuning. It can be seen that not all variables are reported, revealing significant information gaps. For the dataset characteristics (Figure~\ref{fig:variables-percentages-dataset-cleaning-training}(a)), only 22\% of the studies share the datasets used, while its basic statistical details such as the number of samples, mean, and portion of missing data, are reported in just 31\% of the articles. These findings indicate that the preliminary requirements for reproducibility are largely missing in the reviewed articles. Despite these shortcomings, a noteable 80\% of the articles provided meta-data about the datasets. This includes information related to environmental context and data collection details such as time duration, building size, number of floors, number of zones, and, in the case of experimental datasets, descriptions of the laboratory setup and other relevant information. Interestingly, this outcome aligns with findings by~\citep{gundersen2018state}, where about 49\% of articles reported data-related information—the most frequently documented variable. However, our formulation is more granular, while theirs abstract, framed as: \textit{``How well is the dataset documented?"}. In our case, we believe the reason behind the relatively high level of information sharing of this variable, compared to others, lies in the primary research background of the authors. Researchers and engineers in the building energy domain often focus on development model-, heuristics- and knowledge-based techniques, which require an in-depth understanding of building systems and their operational environments. As a result, they tend to inherently provide more comprehensive metadata in such cases. 

\begin{figure}[htbp]
    \centering
    \includegraphics[clip, trim=0.0cm 0.0cm 0cm 0.0cm, width=1\textwidth]{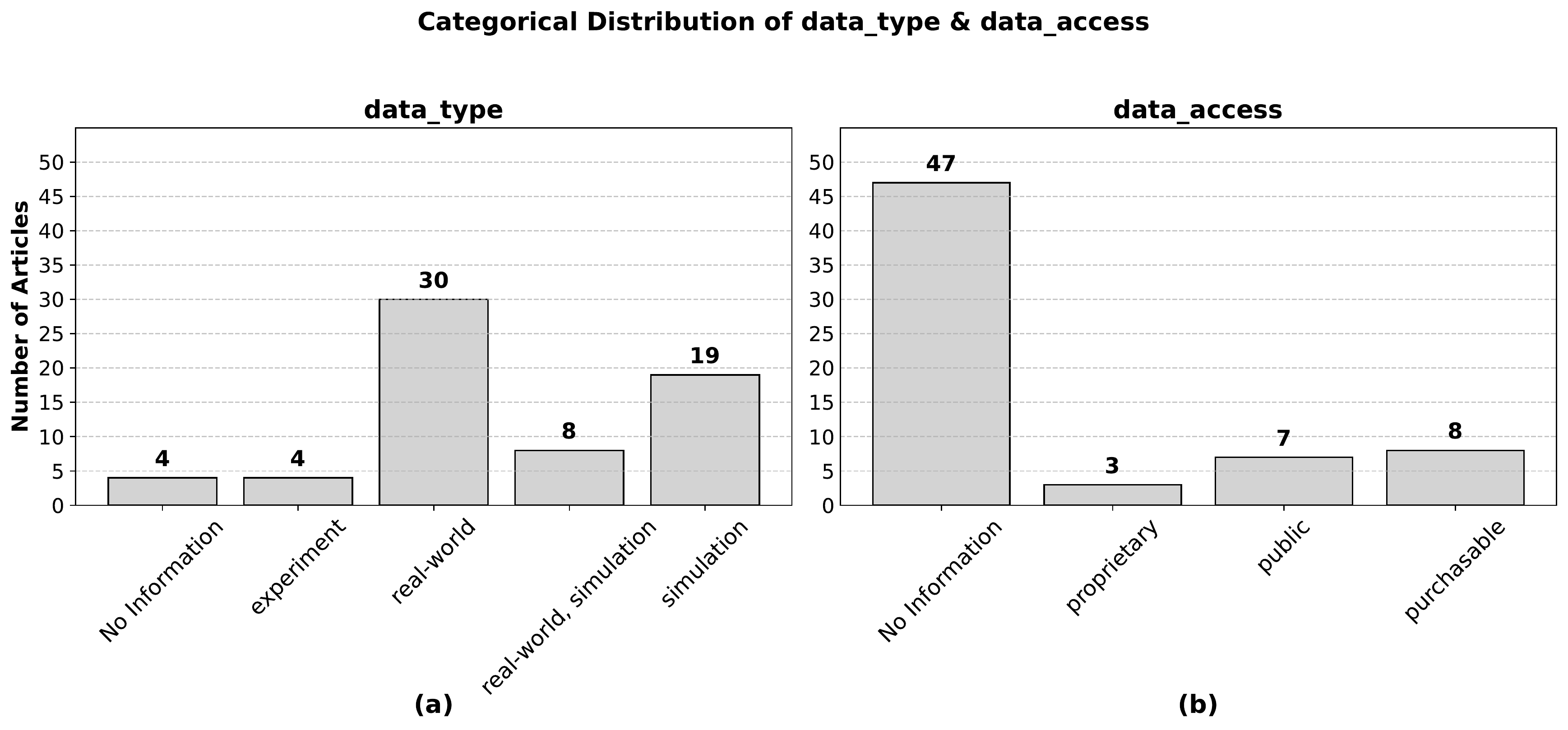}
    \caption{Data Type and Data Accessibility Analysis}
    \label{fig:variables-categorization-dataset-type-access}
\end{figure}

A closer examination combined with the analysis of data type and accessibility (see Figure~\ref{fig:variables-categorization-dataset-type-access}(b)) reveals that approximately 72\% of the articles do not provide any information regarding the nature of the dataset, i.e., whether it is public, proprietary, or commercially available. However, the majority of articles utilize real-world datasets (46\%) as reported in Figure~\ref{fig:variables-categorization-dataset-type-access}. This is a concerning outcome, as one might reasonably assume that real-world datasets are often not made publicly available due to NDAs and data sharing regulations. However, the majority of the articles fail to mention such constraints in their description or sharing of information on how to access the dataset. Alternatively, the absence of any statement regarding dataset accessibility might itself be interpreted as an implicit indication of proprietary constraints—although, in the absence of clarification, one can only assume. Furthermore, we found that nearly half of the articles reported training and testing their methodologies on real-world datasets. This may also reflect the presence of contractual agreements or NDAs with organizations or companies, which may have restricted authors from disclosing detailed information. Nevertheless, it is notable that more than half of the articles make no mention of dataset accessibility or related guidelines. Moreover, Figure~\ref{fig:variables-categorization-dataset-type-access} indicates that the reliance on simulated datasets is reported in 29\% of the articles, followed by the combined use of both real-world and simulation data in 12\%. This type of setting is commonly observed in studies where the methodology is trained on simulated data and tested on real-world datasets. Finally, 6\% of the articles report using experimental datasets, while a similar proportion provide no information regarding the type of dataset used.

Looking at Figure~\ref{fig:variables-percentages-dataset-cleaning-training}(b), data preprocessing related information is reported in only 28\% of the articles, whereas feature generation and their representation are documented in 65\%. We observed consistent use of basic feature preprocessing techniques such as MinMax Scaler and Standard Scaler. For feature representation, we found that articles typically provided input features along with their descriptions, either in tabular form or within the methodology section. Nevertheless, the disparity between these two variables highlights a concerning trend: although many studies emphasize how raw data is transformed into feature representation and input vectors, the initial and equally important part of cleaning and preparing raw data is often under reported. 

Figure~\ref{fig:variables-percentages-dataset-cleaning-training}(c) provides insights into the model training and validation phase, one of the most critical steps in the machine learning model development process. Unfortunately, despite its importance, this category scores the lowest on average among all reproducibility aspects addressed in RQ1. While 32\% of the reviewed articles mention the use of optimization ($opt_{mentioned}$) only 25\% explain the optimization procedure ($opt_{procedure}$), and a mere 5\% report how the baseline model was optimized ($opt_{baseline}$). Even more concerning, only 12\% of articles document model hyperparameters ($params_{model}$), 31\% specify the parameters of the best-performing model ($params_{best\ model}$), and just 3\% provide the corresponding information for the best baseline model ($params_{best\ baseline}$). Optimization procedures, such as grid search, random search, or Bayesian optimization, directly influence model performance. Without access to the optimization setup, including the search space, objective function, and validation strategy, it is difficult to reproduce claimed results or assess whether improvements over baselines are statistically or practically significant~\citep{ferrari2019we}. Strikingly, no article reports the parameter settings for the baseline model ($params_{baseline}$), which is critical for fair comparisons. The absence of details on baseline tuning does not help, as poorly tuned baselines can give a misleading impression of the proposed model's effectiveness~\citep{ferrari2019we}.
\begin{tcolorbox}[
    colback=white,
    title=\textbf{RQ2},
    colframe=gray!40,
    fonttitle=\bfseries,
    coltitle=black,
    boxrule=0.3mm,
    arc=2mm,
    left=2mm,
    right=2mm,
    top=1mm,
    bottom=1mm,
    enhanced
]
The literature significantly underreports the sharing of resources such as coding scripts, datasets, and trained models via external links.
\end{tcolorbox}
Documented information related to code scripts, datasets, and associated executable files is notably scarce. Such resources are typically shared via external links to platforms like Github, Zenodo, or Dropbox. Referring back to Figure~\ref{fig:heatmap-variables}, here we focus on the variable $code_{link}$, which serves as a predicate for the variables positioned to its right in the figure. In essence, the availability of code is a basic indicator that correlates with the reporting of additional implementation-specific details related to feature generation, preprocessing, evaluation and general information about the code. It is evident that variables in this category are completely unreported across the reviewed articles except for two instances where authors provided link to Github repositories (3\%). In one of these cases, the article included a link to Jupyter notebook, but the link was broken and the server returned ``Not Found" error. To recover the resource, we performed an additional investigation, identified the author's Github username, and subsequently located the relevant repository. 

Positively, we found that both articles shared evaluation code including data splitting strategies, input vectors, comparisons with baseline models, and metrics used to estimate the performance gains. This level of detail also extended, in general, to the search ranges used for model fine-tuning. However, in one case, the code appeared to rely on a predefined set of values for training, with varying lagged values for input and output sequences. This may have been an oversight, possibly due to the author forgetting the updated or push the final version of the scrip to the repository prior to publication. The baselines consisted of relatively simple models such as Random Forest and Isolation Forest, and in other case where performance on one building was used as a baseline to compare results on another building, an approach called transfer learning, and appeared to be implemented by the authors themselves. However, when authors implement baselines on their own, particularly when comparing their method against results from prior research or state-of-the-art (SOTA) methods, there is a risk that these implementations may not be entirely accurate. This can potentially lead to misleading comparisons and incorrect conclusions~\citep{hidasi2023effect}. The description in the code repositories ($code_{info}$) is generally limited or minimal. However, we still marked this variable as 1, as some level of information is present.

\begin{tcolorbox}[
    colback=white,
    title=\textbf{RQ3},
    colframe=gray!40,
    fonttitle=\bfseries,
    coltitle=black,
    boxrule=0.3mm,
    arc=2mm,
    left=2mm,
    right=2mm,
    top=1mm,
    bottom=1mm,
    enhanced
]
Information related to data splitting strategies and evaluation metrics is generally well documented across the reviewed articles.
\end{tcolorbox}

\begin{figure}[htbp]
    \centering
    \includegraphics[clip, trim=0.0cm 0.0cm 0cm 0.0cm, width=0.9\textwidth]{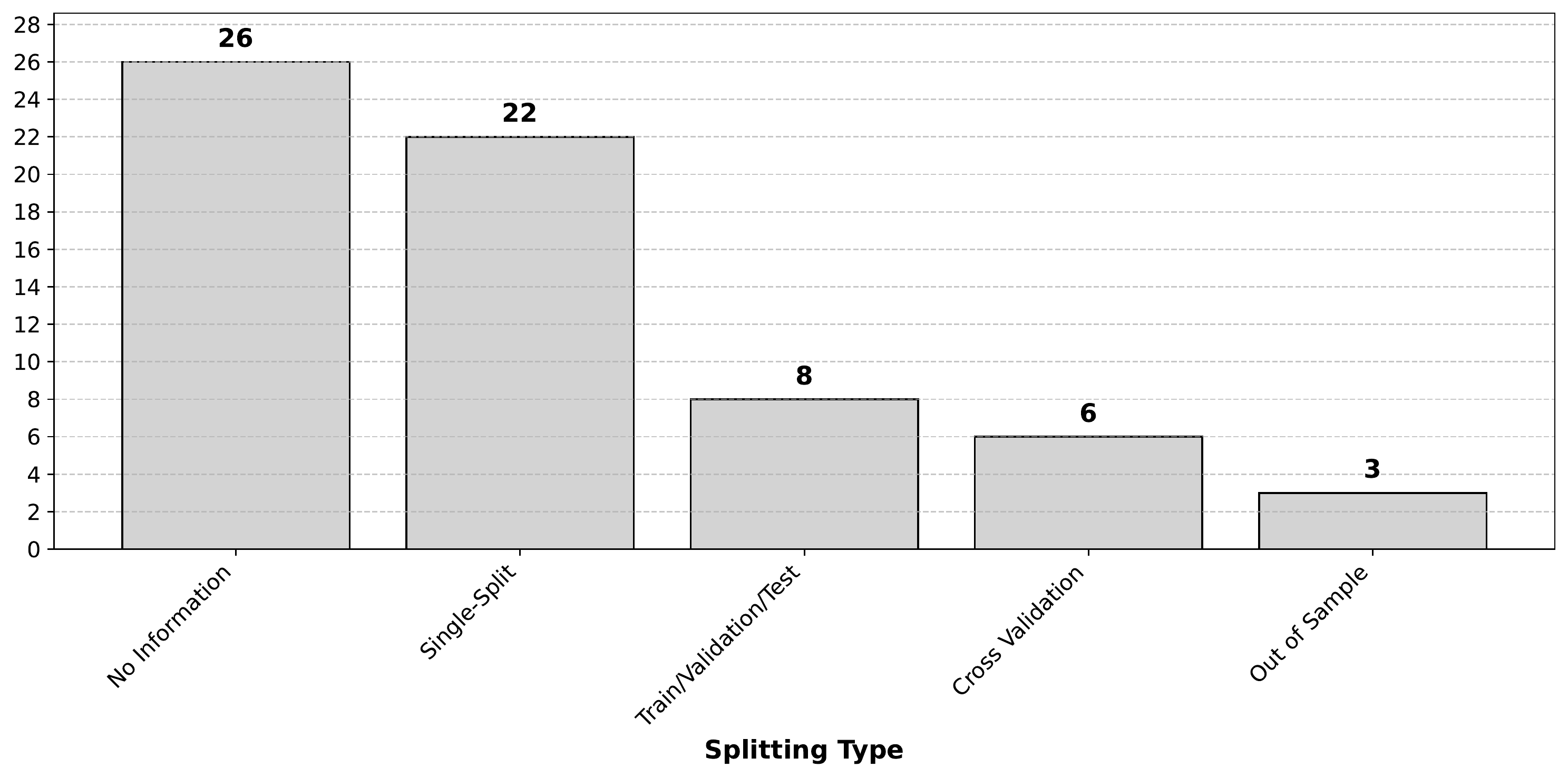}
    \caption{Types of Evaluation Strategies}
    \label{fig:eval_splitting_types}
\end{figure}
Overall, in response to RQ3, we found that approximately 60\% of the articles reported the type of data splitting ($eval_{splitting}$) technique used, with an even higher proportion of 91\% documenting the evaluation metrics ($eval_{mterics}$) used. While data splitting strategies are critical for evaluating model performance, their documentation remains incomplete across many reviewied articles. Looking at Figure~\ref{fig:eval_splitting_types}, which presents the distribution of methodology types, the most frequently reported startegy is single-split validation (34\%). This approach, particularly based on random splits, is generally considered suboptimal and not robust against concept drift and other temporal effects~\citep{lyu2021empirical}. However, in our case, the datasets primarily exhibit time-series characteristics where temporal dependencies exist between observation. Even in such settings, it is important to perform evaluation using walk-forward methodologies or time-based cross-validation techniques to ensure that the model's performance is robust, generalizes well to unseen future instances and avoid the influence of NI-factors~\citep{pham2020problems}. Unfortunately, we found a relatively low rate of reporting cross-validation procedures, only 9\% of the articles documented their use. Similarly, 12\% of the articles reported using a Train/Test/Validation split, a suitable technique re-adjusting parameter weights during model optimization is necessary. Such a low rate is also consistent with earlier observations indicating limited reporting of hyperparameter fine-tuning details. Furthermore, only 5\% of the articles employed out-of-sample evaluation methods, which involve testing the model on data that was not used during either the training or validation phases. This strategy is particularly useful for assessing model performance on truly unseen data, essentially, its ability to \emph{generalize to an external population}~\citep{varoquaux2023evaluating}. However, this technique is not commonly reported in the reviewed articles. 
\begin{figure}[htbp]
    \centering
    \includegraphics[clip, trim=0.0cm 0.0cm 0cm 0.0cm, width=0.8\textwidth]{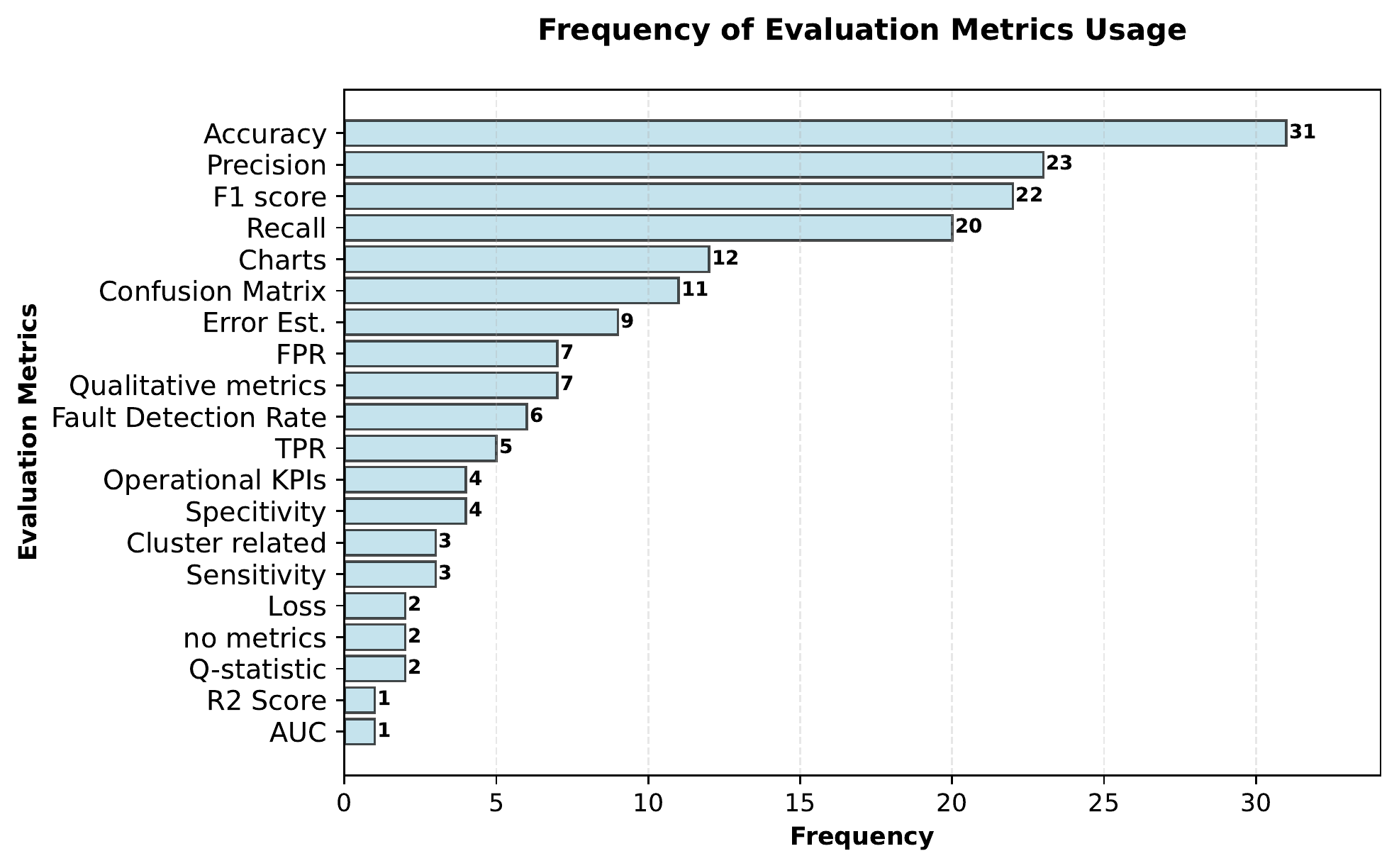}
    \caption{Frequency of Evaluation Metrics}
    \label{fig:eval_metrics_freq}
\end{figure}

Evaluation metrics are among the most consistently reported indicators in the reviewed articles, with nearly 91\% of the studies including them. Figure~\ref{fig:eval_metrics_freq} presents the metrics used, ordered by the frequency of occurenece. A clear cut-off appears after the first four metrics, Accuracy, Precision, F1 Score, and Recall, which are the most prevalent in FDD studies and are consistently reported accort the reviewed literature. In addition, other quantitative metrics such as False Positive Rate (FPR), True Positive Rate (TPR), Error Estimation, Fault Detection Rate, and Confusion Matrix are also commonly used. In some cases, qualitative metrics and operational key performance indicators (KPIs) relevent to building systems are reported, for example, reduction in man-hours, or time elapsed between fault detection and diagnosis or performance monitoring through visualization/charts. In addition, significance testing ($sig_{test}$; see Figure~\ref{fig:heatmap-variables}) to assess performance gains over appears in only one article. Interestingly, some articles include this test as part of the fault detection methodology; however, this measure typically do not appear in the evaluation phase and remain significantly underreported. Significance testing is common in machine learning scientific community, for instance, the Friedman test often supports comparisons of multiple classifiers across multiple datasets~\citep{demvsar2006statistical}.

\subsection{Assessment of Reproducibility Degree}
\label{subsec:results-assessment-reprod-dims}
\begin{figure}
    \centering
    \includegraphics[width=0.75\textwidth]{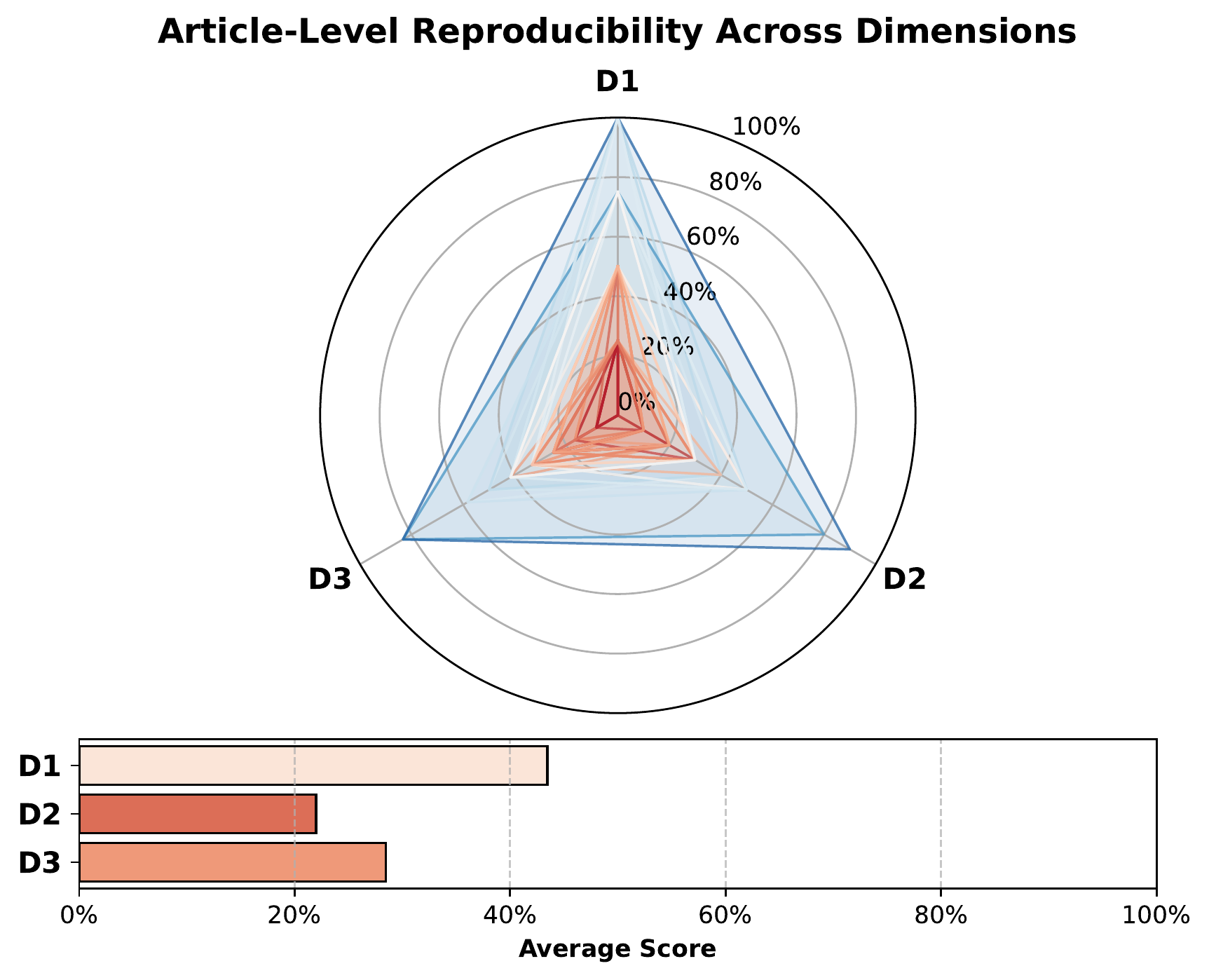}
    \caption{Dimension Scores}
    \label{fig:spider-web-dimensions}
\end{figure}

In the previous section, we presented detailed insights into the documentation of reproducibility-related variables. The results are particularly concerning for aspects related to model training and the availability of code repositories, where reporting appears limited or, in some cases, significantly underreported. As a next step, we aim to anlayze reproducibility across thee key dimensions: $D_{1}$ (Data), $D_{2}$ (Methodology), and $D_{3}$ (Experiment). This analysis provides a more comprehesive view of how well individual studies support reproducibility crucial components. In this context, Figure~\ref{fig:spider-web-dimensions} presents the quantified assessment of each dimension for the reviewed studies. In most cases, the concentration of measured scores appears near the center of the plot along each axis, on average ranging between 5\% and 25\%, indicating that the majority of articles fail to report variables more often. A few cases can be easily spotted where the scores are comparatively higher, particularly for dimension $D_{1}$. In total, six articles document all the variables associated with dimension $D_{1}$. However, only two articles achieve an overall reproducibility score of approximately 80\% for $D_{2}$ and $D_{3}$, respectively. This observation may suggest that a few authors inherently adopt more detailed documentation practices, or that certain reporting standards align with requirements imposed by funding agencies or institutional policies.

While the overall scores across all three dimensions (as shown in the subplot) remain relatively low, ranging between 22\% and 44\%, this indicates that nearly two-thirds of the reproducibility variables are not documented in the reviewed articles, particularly within dimensions $D_{2}$ and $D_{3}$.~A general trend is that dataset characteristics are reported more frequently; however, this coverage remains insufficient, with over 50\% of the variables in this category still undocumented. Although dataset information appears to be the clear winner here, this may reflect the disciplinary background of the authors, many of whom are likely more aligned with building science or industry practice and may not be as engaged with the machine learning research community. Another way to look at it is that $D_{1}$ includes fewer variables than $D_{2}$ and $D3$, and some of those, like $data_{metadata}$, are less granular. In contrast, the variables in $D_{2}$ and $D_{3}$ have a more detailed and specific to the development of ML methodology, making their absence more noticeable during the review. Further, the assumption underlying this quantification is the independence among the variables. It represents the normalized aggregated sum over the variables for each category. In other words, if the data sources are not listed, the outcome reflects a partial or missing contribution to the overall reproducibility score for the data dimension, regardless of the completeness of other variables in that category. 

\begin{figure}
    \centering
    \includegraphics[width=0.6\textwidth]{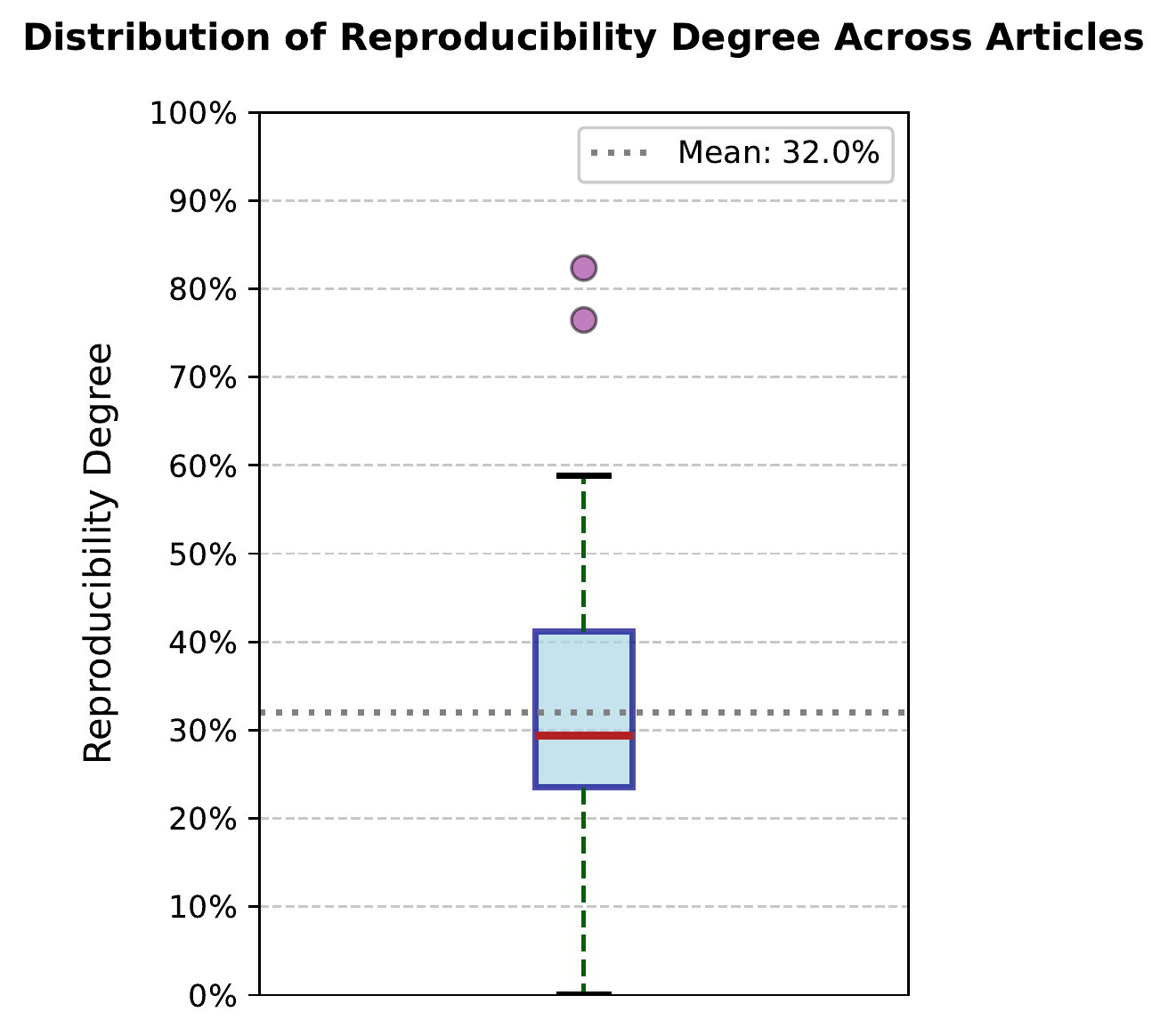}
    \caption{Reproducibility Degree of Articles}
    \label{fig:reprod-degree-articles}
\end{figure}
The overall assessment of reproducibility degree, computed according to Equation~\ref{equ:degree-calc}, is presented in Figure~\ref{fig:reprod-degree-articles}. The objective here is to evaluate the extent of information shared across all three dimensions. To this end, the reproducibility variables from dimension $D_{1}$, $D_{2}$, and $D_{3}$ are combined, and a normalized score, referred to as the overall reproducibility degree, is assigned to each article. As shown in the figure, the average reproducibility degree is 32\%, i.e., on average articles report only about one-third of the information that is crucial to reproduce their experiments. It is important to note that this metric does not require all variables to be reported (i.e., conjunction); otherwise, the score would be zero for all articles. This is a concerning results, as it reflects a broader systemic issue. To put this in perspective, it suggests that articles published over the past decade offer limited information to support reproducibility, despite growing awareness of its importance in scientific research and the ongoing debate surrounding it~\citep{fidler2018reproducibility, haibe2020transparency, national2019reproducibility}.

\subsection{Scientometric Analysis }
\label{subsec:Scientometric Analysis }
We would like to add another nuance to the discussion by taking into account both the institutional background of the authors—specifically, whether they are affiliated with academic or industrial institutions—and the subject areas in which they work. We consider this distinction important, as there may be legitimate reasons why publications originating from industrial contexts follow different reproducibility practices. A prominent example is the withholding of code or data due to intellectual property concerns. In addition, subject areas—such as computer science or applied engineering—may shape the norms and expectations surrounding reproducibility.

To evaluate whether authors are affiliated with academic or industrial institutions, we classified each author into one of three categories: university, industry, or non-academic research institutions (e.g., public research centers or government laboratories). Since some authors reported multiple affiliations, the total number of affiliations exceeds the number of individual authors. In total, we identified 210 university affiliations, 26 non-academic research affiliations, and 36 industry affiliations. At the paper level, 43 of the 65 publications were authored exclusively by university-affiliated researchers. In contrast, only 5 papers were authored solely by researchers from industry, and 1 paper solely by authors from non-academic research institutions. The remaining publications were authored by teams with mixed institutional backgrounds.

To analyze the subject areas associated with the authors, we retrieved metadata for all articles using the Scopus API~\citep{scopusapi}. For each article, we accessed the Scopus Author Profiles of the listed authors and extracted the subject areas associated with each profile. Scopus assigns up to three subject areas per author, based on their publication history, which reflect their primary fields of research. These classifications allow us to identify the disciplinary contexts in which the authors primarily conduct their research. In total, 173 authors were associated with Engineering as a subject area, followed by 143 with Computer Science, 46 with Energy, 36 with Mathematics, 17 with Environmental Science, 12 with Physics and Astronomy, and 10 with Decision Sciences. All other subject areas were assigned to fewer than ten authors. At the paper level, 60 of the 65 papers included at least one author associated with Engineering, and 55 papers included at least one author associated with Computer Science. 

We explored potential associations between (i) reproducibility scores and citation counts, (ii) reproducibility scores and institutional affiliation, and (iii) reproducibility scores and whether at least one author is associated with Computer Science as a subject area. Visual inspection of the data (see plots in the GitHub repository) did not suggest any obvious differences. We emphasize that the dataset comprises only 65 papers, which limits the statistical power of any formal analysis. 


\section{Discussion}  
\label{sec:discussion}
Over the past decade, the term \emph{reproducibility crisis} has become ubiquitous across scientific disciplines, indicating growing concern over the reliability of published results~\citep{baker20161, gundersen2018state, hutson2018artificial, fidler2018reproducibility}. 
In the literature, factors contributing to irreproducible results have been highlighted, including the incentive culture of “publish or perish,” competition for research funding, poorly designed statistical methods, and inadequate publishing practices—issues particularly prevalent in the behavioral and life sciences ~\citep{fidler2018reproducibility, cobey2024biomedical, schweiger2024costs}. 
A widely acknowledged barrier to reproducibility in ML-based research is the lack of access to data.
There are legitimate reasons why authors cannot share original data. 
Authors—regardless of whether they are affiliated with academia or industry—may face barriers to data sharing and transparency due to privacy or ethical restrictions. Additionally, industrial authors may be constrained by intellectual property concerns or competitive constraints.
However, the literature discusses various strategies to address these barriers. Even when original datasets cannot be made publicly available, authors could, for instance, provide synthetic datasets that replicate the structure and statistical properties of the real data while preserving privacy ~\citep{semmelrock2025reproducibility}.
Another widely discussed issue in the context of reproducibility is the limited availability of code. As with data, there may be legitimate reasons for withholding code—particularly for industry-affiliated authors, who may be bound by intellectual property or competitive constraints. However, it becomes increasingly difficult to justify the absence of code sharing in publications authored exclusively by academic researchers. In our sample, this applied to 43 of the 65 analyzed papers. Even when code is made available, it is often incomplete or poorly documented, which limits its utility for reproduction and reuse ~\citep{trisovic2022large, cremonesi2021progress}.
There are several potential reasons for this. A recent study surveyed scientists from various disciplines to investigate their programming backgrounds, practices, and the challenges they face regarding code readability ~\citep{chen2025exploring}. The findings indicate that most participants learned programming through self-study or on-the-job training, with limited formal education in writing readable code. In addition, the study found low adoption of code quality tools and a growing trend toward using large language models to improve code quality.

~\cite{semmelrock2025reproducibility} discussed potential drivers toward improving ML reproducibility.
First, raising awareness of reproducibility challenges and providing targeted training or educational resources can support efforts to improve reproducibility across disciplines ~\citep{wiggins2019replication}.
Second, policies implemented by journals and conferences play a crucial role in promoting transparency and setting minimum expectations for sharing code and data. These institutions can build on existing checklists and reproducibility guidelines to formalize and enforce best practices ~\citep{pineau2021improving}.
Third, the use of privacy-preserving technologies can enable the utilization of sensitive data while allowing data owners to collaboratively train models on private datasets.

\section{Conclusion}  
\label{sec:conclusion}
Reproducibility allows for independent verification of results, facilitates the identification of errors, and supports the cumulative advancement of knowledge.
In this work we undertake a comprehensive assessment of the reproducibility machine learning-based fault detection and diagnosis in building energy systems. 
We curated a checklist of reproducibility variables grounded in established literature and extended it specifically for ML-based FDD applications in building energy systems. 
These variables were grouped into three reproducibility dimensions—data, methodology, and experiment—to systematically evaluate the degree to which studies document critical information.
None of the articles can be fully reproduced. 
To assess the degree of reproducibility, each dimension was assigned a reproducibility score, reflecting the extent to which information and materials are shared to support reproducibility. 
On average, the data dimension scored 43\%, the methodology dimension 22\%, and the experiment dimension 28\%. 
A widely acknowledged barrier to reproducibility is the lack of access to both code and data.
72\% of the articles do not specify whether the dataset used is public, proprietary, or commercially available. 
Only two papers share a link to their code—one of which was broken. 
We explored potential associations between reproducibility scores and factors such as citation counts, institutional affiliation (academic vs. industrial), and whether at least one author is associated with computer science as a subject area in their Scopus author profile.
No differences in reproducibility were observed. 
While there may be legitimate reasons why publications originating from industrial contexts adhere to different standards or conventions regarding the documentation and sharing of research artifacts, our analysis shows that two-thirds of the studies were authored exclusively by academic researchers—yet their reproducibility scores did not differ from those of studies involving industry-affiliated authors.
These findings highlight the need for targeted interventions, including reproducibility guidelines, training for researchers,
and policies by journals and conferences that promote transparency and reproducibility.

A list of the reviewed articles, the recorded reproducibility variables, and all analysis scripts and data are openly available at: \url{https://github.com/tuw-isab/reproducibility-analysis-ml-based-fdd-hvac}.

\section*{Acknowledgements}
The research leading to these results was conducted within the project Ecom4Future (project number 903927), funded by the Austrian Research Promotion Agency (FFG) under the “Clean Energy Transition Partnership – Joint Call 2022”.
\bibliographystyle{elsarticle-harv}

\bibliography{references}
\end{document}